\documentclass[sigconf]{acmart}

\AtBeginDocument{%
  }

\begin{document}

\title{CookingDiffusion: Cooking Procedural Image Generation with Stable Diffusion}

\author{Yuan Wang}
\affiliation{%
  \institution{University of Science and Technology of China}
  \city{Hefei}
  \state{Anhui}
  \country{China}
}
\email{wy1001@mail.ustc.edu.cn}

\author{Bin Zhu}
\email{binzhu@smu.edu.sg}
\affiliation{%
  \institution{Singapore Management University}
  \city{Singapore}
  \country{Singapore}
}

\author{Yanbin Hao}
\affiliation{%
    \institution{Hefei University of Technology}
  \city{Hefei}
  \state{Anhui}
  \country{China}}
\email{haoyanbin@hotmail.com}

\author{Chong-Wah Ngo}
\affiliation{%
\institution{Singapore Management University}
  \city{Singapore}
  \country{Singapore}
}
\email{cwngo@smu.edu.sg}

\author{Yi Tan}
\affiliation{%
 \institution{University of Science and Technology of China}
  \city{Hefei}
  \state{Anhui}
  \country{China}}
\email{ty133@mail.ustc.edu.cn}

\author{Xiang Wang}
\affiliation{%
  \institution{University of Science and Technology of China}
  \city{Hefei}
  \state{Anhui}
  \country{China}}
\email{xiangwang1223@gmail.com}

\begin{abstract}
Recent advancements in text-to-image generation models have excelled in creating diverse and realistic images. This success extends to food imagery, where various conditional inputs like cooking styles, ingredients, and recipes are utilized. However, a yet-unexplored challenge is generating a sequence of procedural images based on cooking steps from a recipe. This could enhance the cooking experience with visual guidance and possibly lead to an intelligent cooking simulation system. To fill this gap, we introduce a novel task called \textbf{cooking procedural image generation}. This task is inherently demanding, as it strives to create photo-realistic images that align with cooking steps while preserving sequential consistency. To collectively tackle these challenges, we present \textbf{CookingDiffusion}, a novel approach that leverages Stable Diffusion and three innovative Memory Nets to model procedural prompts. These prompts encompass text prompts (representing cooking steps), image prompts (corresponding to cooking images), and multi-modal prompts (mixing cooking steps and images), ensuring the consistent generation of cooking procedural images. To validate the effectiveness of our approach, we preprocess the YouCookII dataset, establishing a new benchmark. Our experimental results demonstrate that our model excels at generating high-quality cooking procedural images with remarkable consistency across sequential cooking steps, as measured by both the FID and the proposed Average Procedure Consistency metrics. Furthermore, CookingDiffusion demonstrates the ability to manipulate ingredients and cooking methods in a recipe. We will make our code, models, and dataset publicly accessible.
\end{abstract}




\settopmatter{printacmref=false}
\setcopyright{none}
\renewcommand\footnotetextcopyrightpermission[1]{}
\pagestyle{plain}

\maketitle

\section{Introduction}

Have you ever been frustrated when you cannot understand a recipe during cooking? Do you feel more delighted with the recipe along with image guidance for each step when cooking? Have you ever imagined customizing your personal recipes with an intelligent cooking simulation system? To achieve these goals, as shown in Fig.~\ref{introduction} (c), we propose a novel task named cooking procedural image generation. Its objective is to generate a sequence of consistent images, each aligning with a specific step in a recipe. 
\definecolor{myyellow}{RGB}{255, 220, 127}
\definecolor{myblue}{RGB}{145, 172, 224}
\label{sec:intro}
\begin{figure*}[t]
\centering
\includegraphics[width=\textwidth]{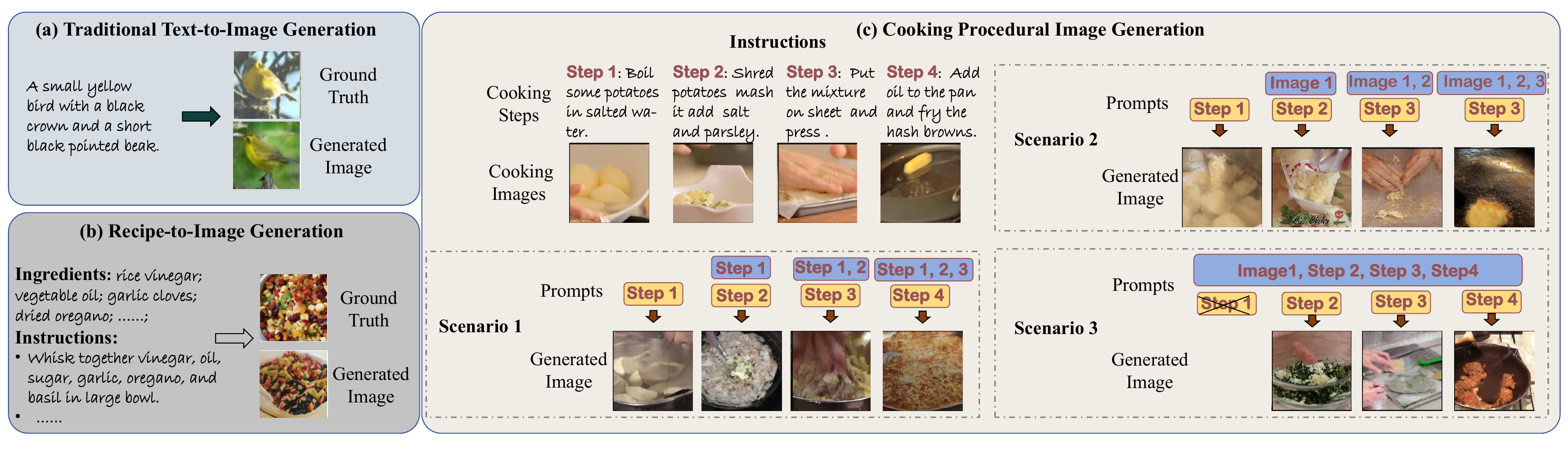}
\caption{Tasks comparison of traditional text-to-image generation, recipe-to-image generation, and our proposed cooking procedural image generation. (a) Traditional text-to-image generation involves generating an image based on a textual description. (b) Recipe-to-image generation aims to generate the final dish image based on the entire recipe. (c) Our proposed cooking procedural image generation task aims to generate a sequence of consistent cooking images that correspond to each specific step in a recipe, where the current step (\textcolor{myyellow}{yellow blocks}) is regarded as the conditional prompt and the previous contextual steps (\textcolor{myblue}{blue blocks}) as procedural prompts.} 
\Description{An illustration of the difference between the proposed cooking procedural image generation task and current generative tasks.}
\label{introduction}
\end{figure*}

Great progress has been achieved in recent years for text-to-image generation~\cite{rombach2022high,ho2020denoising,mirza2014conditional, reed2016generating2, zhang2018stackgan++, zhang2017stackgan}. As shown in Fig.~\ref{introduction} (a), the goal of traditional text-to-image generation is to generate an image aligned with the text description, following a principle involuntary where the generated images are solely influenced by the corresponding text prompts. In other words, the generated images are considered independent of each other. 
In the food domain, recipe-to-image generation is explored in CookGAN \cite{zhu2020cookgan}, which generates the final dish images based on the entire recipe including ingredients and the text instruction shown in Fig.~\ref{introduction} (b). In contrast, in this paper, we focus on generating cooking images for each step within a recipe, i.e., a procedure. Note our cooking procedural image generation task is different from text-to-video generation \cite{ho2022video, yin2023nuwa, yu2023video, villegas2022phenaki, ho2022imagen}. Video generation models emphasize rigid temporal continuity, where the variations between adjacent frames are minimal, and the visual content mostly depicts a single scene with minor changes in the background. In contrast, our cooking procedural image generation task involves diverse scene variations and prioritizes consistency rather than rigid temporal continuity. For instance, consider steps 2-4 in Fig.~\ref{introduction} (c) [``\textit{Shred potatoes mash it add salt and parsley.}'', `` \textit{Put the mixture on sheet and press.}'', ``\textit{Add oil to the pan and fry the hash browns}''], where adjacent steps vary in terms of background and ingredients state. 

The challenges of cooking procedural image generation lie in three folds. \textbf{First}, it is non-trivial to maintain the consistency of the generated procedural images in sequential order. 
Consider two adjacent steps in Fig.~\ref{introduction} (c): [``\textit{Shred potatoes mash it add salt and parsley.}'', ``\textit{Put the mixture on sheet and press.}''], traditional text-to-image generation models would treat these steps as two individual prompts and generate corresponding images independently. However, in this case, accurately interpreting the pronoun ``\textit{mixture}'' in the second step and ensuring consistency between the steps becomes challenging for these models. %
Additionally, it is also necessary to maintain consistency in elements such as the background, cooking utensils, and ingredients throughout the generated images of a recipe. This necessitates considering not only the current step text but also the contextual information within the recipe to incorporate the relationships among the step texts in the generation process. \textbf{Second}, the procedural prompts for cooking procedural image generation can go beyond text prompts based on cooking steps, such as image prompts and even multi-modal prompts to capture contextual information. As shown in Fig.~\ref{introduction} (c), we consider three scenarios with different procedural prompts, which aim to guide the generation process to maintain consistency. This is achieved by considering the current cooking step as the conditional prompt while treating the previous contextual steps as procedural prompts.
\begin{itemize}
\item \textbf{Scenario 1} involves only text prompts, specifically cooking step descriptions. Procedural prompts include all previous cooking steps. For instance, procedural prompts for generating the image of step 4 are steps 1-3 while the text of step 4 is used as a conditional prompt.
\item \textbf{Scenario 2} employs only cooking images as procedural prompts, utilizing corresponding images for each historical step to offer valuable contextual information. For example, generating the image for step 4 involves combining cooking images from steps 1-3. 
\item \textbf{Scenario 3} introduces multi-modal procedural prompts by combining step-wise texts and images. Leveraging the available cooking images along with cooking instructions for steps lacking images, the objective is to generate images for those steps without cooking images. Recognizing that users may not upload images for every cooking step, this scenario would pave the way to generate the missing images for all the steps, thereby enhancing the cooking experience with comprehensive visual cues.
\end{itemize} 
In addition, the \textbf{third} challenge lies in dealing with the cause-and-effect in cooking steps for image generation, which is commonly observed in recipe-to-image generation \cite{zhu2020cookgan}. This challenge entails not only emphasizing the cooking actions but also accurately representing the outcomes of each cooking step.

To address these issues, we propose CookingDiffusion, which incorporates three novel Memory Nets into Stable Diffusion \cite{rombach2022high}. In CookingDiffusion, the current step text serves as a necessary conditional prompt for generating the desired image. These Memory Nets are tailored to handle various types of procedural prompts, enhancing Stable Diffusion's capability to generate high-quality and consistent procedural images. Specifically, the first two types of Memory Nets are Text Memory and Image Memory Nets, which are introduced to deal with the above-mentioned scenarios 1 and 2 to learn text-based and image-based procedural representations, respectively.
The Text Memory Net is tailored for inputs with only procedural text prompts, learning text-based procedural representations and incorporating them into the generation process. On the other hand, the Image Memory Net is designed for image-based procedural representations, ensuring the effectiveness of visual cues in maintaining consistency during image generation. Both of these Memory Nets are simple yet very effective in learning and incorporating procedural representation from both textual and visual modalities. It is noteworthy that both modules share a unified structure for processing procedural prompts across diverse modalities. Moreover, this unified structural framework proves invaluable in addressing the third scenario. Multi-modality Memory Net is introduced to fuse the Text Memory Net and Image Memory Net so that it is capable of processing multi-modal procedural prompts.

As none of the existing datasets could be directly used for our cooking procedural image generation task, we pre-process YouCookII dataset~\cite{zhou2018towards} for training and evaluating our CookingDiffusion. Based on the available annotations and timestamps of each step, we segment the videos in YouCookII and select keyframes as ground truth step images using CLIP scores~\cite{radford2021learning}. In addition, we introduce the Average Procedure Consistency (Avg-PCon) metric based on the CLIP score, to evaluate the consistency among the generated procedural images. Avg-PCon quantifies the consistency within a procedure by calculating the CLIP score between the generated images and the texts that do not correspond to them within a recipe.

Our contribution is summarized as follows: (a) Pioneering the definition of a novel task, cooking procedural image generation, encompassing three distinct prompt scenarios. (b) Proposing CookingDiffusion, featuring three innovative memory nets to handle diverse procedural prompts and ensure high-quality, consistent procedural image generation. (c) Pre-processing the YouCookII dataset to establish a suitable benchmark for our task, which will be publicly accessible. (d) Introducing the Average Procedure Consistency (Avg-PCon) metric to evaluate the consistency among generated procedural images.

\section{Related Work}
\label{sec:related work}
Food computing~\cite{min2019survey, zhao2021fusion, 9179998, yin2023foodlmm} has gained significant research attention recently, covering various tasks like food category recognition~\cite{qiu2022mining, liu2024canteen}, ingredient composition analysis~\cite{chen2016deep, chen2020study, min2023large}, recipe retrieval~\cite{min2016being, salvador2017learning, zhu2019r2gan, zhu2021learning, wang2021cross, salvador2021revamping, song2025enhancing}, recipe generation~\cite{salvador2019inverse, chhikara2024fire, liu2024retrieval}, food logging and dietary tracking~\cite{ming2018food, sahoo2019foodai}. This research is pivotal for health management, addressing diet-related diseases like obesity, diabetes, and cardiovascular conditions. In this paper, we focus on cooking procedural image generation based on cooking steps and images, thus we mainly review the related works in text-to-image generation and food image generation below.

Text-to-image generation is a sub-task of conditional generative tasks \cite{mirza2014conditional}, which was first proposed in \cite{mansimov2015generating}. Given a text description, the generated image should accurately represent the content of the text description. Building upon the success of Generative Adversarial Networks (GANs) \cite{goodfellow2014generative}, GANs have been extended to tackle the text-to-image generation task \cite{mirza2014conditional}. And subsequently, this line of work is further improved to generate vivid and high-resolution images, such as text-conditional convolutional GAN \cite{reed2016generative}, GAWWN \cite{reed2016learning}, StackGAN \cite{zhang2017stackgan}, StackGAN++ \cite{zhang2018stackgan++}, AttnGAN \cite{xu2018attngan}, XMC-GAN \cite{zhang2021cross}, DFGAN \cite{tao2022df}, RiFeGAN2 \cite{9656731} and MSCGAN\cite{zhao2023multi}. However, GAN-based text-to-image generative models often face challenges such as mode collapse and training instability \cite{goodfellow2016nips}. 

Diffusion Models \cite{ho2020denoising}, a likelihood-based generative model, have gained attention for superior image quality and training stability \cite{dhariwal2021diffusion}, but they also encounter challenges. The inference process is computationally intensive, necessitating multiple iterations. To enhance efficiency, fast sampling algorithms \cite{song2020denoising, lu2022dpm, lu2023dpmsolver} have been developed. Additionally, the computational cost of diffusion and reverse processes in image space, particularly for high-resolution images, can be prohibitive. Latent Diffusion Models \cite{rombach2022high} address this computational challenge by using an encoder-decoder architecture based on Variational Autoencoders \cite{van2017neural, esser2021taming}. Stable Diffusion \cite{rombach2022high} extends this concept to large-scale generative models, showcasing robust capabilities that can be applied across a spectrum of generative tasks \cite{10445245}. 
Despite success in text-to-image generation, these models generate images independently for different prompts, adhering to the principle that generated images are associated only with their corresponding conditional prompts. However, this principle is not suitable for our cooking procedural image generation task. Our CookingDiffusion builds on Stable Diffusion, incorporating Memory Net for consistent procedural image generation.

There are also models specifically designed for food image generation tasks, which can be considered as a special scenario within the text-to-image generation domain. In these models, conditional prompts are not always recipes. For example, in \cite{horita2019unseen}, the conditional prompt is the food style, while in \cite{han2019art, ito2018food}, the prompt consists of the ingredients. CookGAN \cite{zhu2020cookgan} is specifically designed to generate food images based on provided cooking steps and ingredients. By addressing the causality effect, CookGAN is capable of generating vivid and high-resolution food images. However, it is important to note that the food image generation task differs from our proposed task. In food image generation, the objective is to generate the final dish images, whereas, in our task, the objective is to generate images corresponding to each step within a recipe. \cite{lu2023multimodal} propose a multimodal procedural planning task that leverages LLMs and multimodal generation models to complement each other in generating procedural plans. However, their approach focuses solely on the content of individual steps and fails to address the consistency across the overall procedure. To the best of our knowledge, cooking procedural image generation is an unexplored problem.

\section{Method}

\begin{figure*}[!t]
\centering
\includegraphics[width=\linewidth]{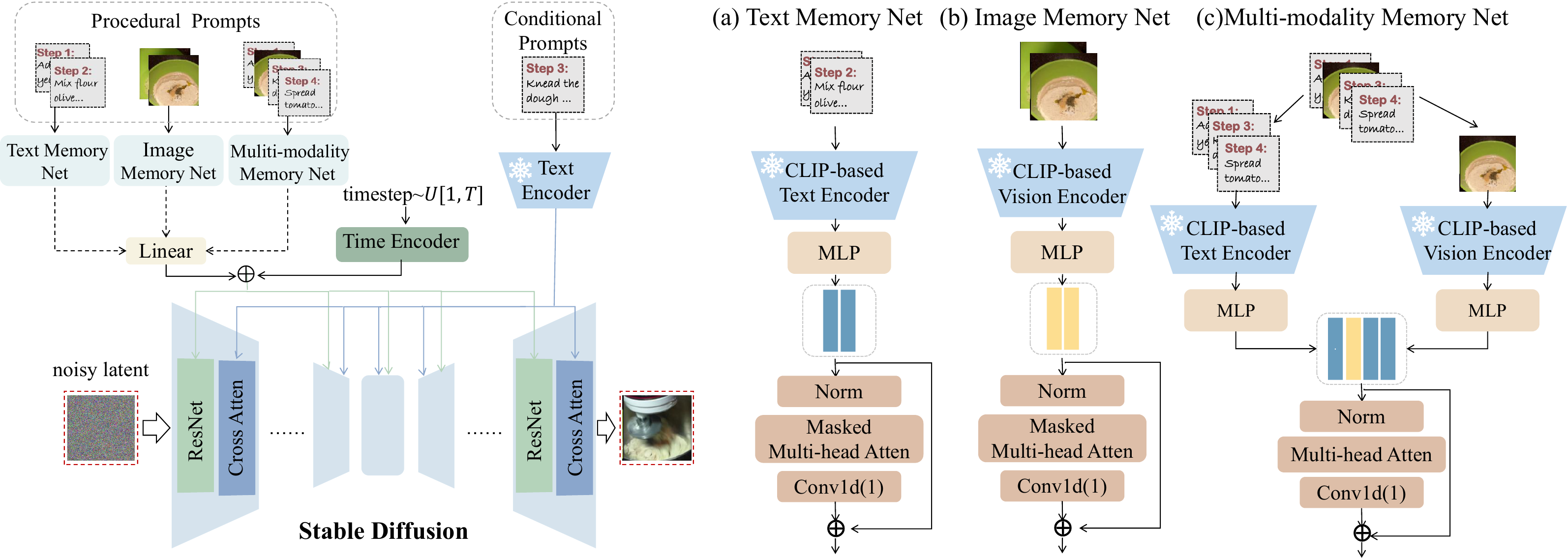}
\caption{Overview of our proposed CookingDiffusion. We introduce three different Memory Nets for CookingDiffusion. (a) The Text Memory Net is tailored for processing text-based procedural prompts, while (b) the Image Memory Net is dedicated to handling image-based procedural prompts. (c) The Multi-modality Memory Net is introduced to deal with procedural prompts from different modalities.}
\label{overview}
\Description{Overview of our proposed CookingDiffusion.}
\end{figure*}

\subsection{Problem Definition}
Given a recipe with multiple cooking steps (i.e., a procedure) as well as procedural prompts in context (e.g., cooking images in history) as input, the goal of cooking procedural image generation is to generate consistent cooking images that correspond to each step. We formulate the three scenarios for our cooking procedural image generation task as follows. 
\begin{itemize}
\item \textbf{Scenario~1}~generates an image sequence denoted as $[\mathbf{I}_{i}^{gen}]_{i=1}^{N}$ solely guided by the text sequence $[\mathbf{C}_{i}]_{i=1}^{N}$, where $\mathbf{C}_{i}$ denotes the text instruction of the $i$-th step, and $N$ is the total number of steps in the recipe. To generate $\mathbf{I}_{i}^{gen}$, $\mathbf{C}_{i}$ is used as the conditional prompt and the historical cooking steps  $[\mathbf{C}_{j}]_{j=1}^{i-1}$ are used as procedural prompts to ensure consistency among the generated images.

\item \textbf{Scenario~2}~utilizes the available image sequence in history, denoted as $[\mathbf{I}_{j}^{known}]_{j=1}^{i-1}$, as procedural prompts. 
In conjunction with the textual conditional prompt $\mathbf{C}_{i}$, the objective remains the same: to generate each step image $\mathbf{I}_{i}^{gen}(i=1,\cdots, N)$ within the entire recipe.

\item \textbf{Scenario~3}~introduces a multi-modal approach where a certain percentage, denoted as $p$, of cooking images is available, and $1-p$ represents the ratio of the used text instruction. This results in a multi-modal procedural prompt sequence, expressed as $[\mathbf{C}_{n_{i}}, \mathbf{I}_{m_{j}}^{known}]; i=1,\cdots, N_{1}, j=1, \cdots, N_{2}$, where $n_{i}$ and $m_j$ represent the positions of the textual and image steps, respectively. The total procedure length remains $N_{1}+N_{2}=N$.
\end{itemize}

\subsection{Model Overview}
As shown in Fig.~\ref{overview}, \textbf{CookingDiffusion} is tailor-made for cooking procedural image generation based on Stable Diffusion. It is apparent that the proposed task cannot be accomplished through Stable Diffusion with only conditional text prompts. This motivates us to introduce procedural prompts to ensure consistency in the generated images. Following the three scenarios mentioned before, we explore three different input settings, i.e., solely text instructions as both conditional and procedural prompts, texts as conditional prompts and images as procedural prompts, and a mix of text prompts for certain steps and image prompts for remaining steps within the recipe as procedural prompts. 

To be specific, procedural prompts are employed to compute what we term ``procedural representation" in an end-to-end manner. The learned procedural representation can be categorized as \textbf{text-based procedural representation}, \textbf{image-based procedural representation}, and \textbf{multi-modal procedural representation} across three scenarios, respectively. It is worth noting that for obtaining image-based procedural representations, we utilize ground-truth step images instead of generated images due to time cost. If generated images are used, the step images could not be generated in parallel and would need to be generated autoregressive, which is extremely time-consuming. Thus we use ground truth images to assess the effectiveness of the procedural prompts learned from visual cues but also provide an upper bound on the performance improvement by image-based procedural prompts. It is a feasible idea in practice, as cooking step images could be available for some recipes on the recipe-sharing website.

To deal with different scenarios of procedural prompts, we propose three kinds of memory networks, named \textbf{Text Memory Net}, \textbf{Image Memory Net}, and \textbf{Multi-modality Memory Net}. Here, the term ``Memory'' emphasizes the historical aspects of procedural texts and images. These Memory Nets are then integrated with Stable Diffusion to facilitate CookingDiffusion. It is commonly accepted that Stable Diffusion necessitates a learnable time encoder for incorporating timesteps into the denoising process. Our Memory Net operates as a parallel module alongside the time encoder. The acquired procedural representations are then combined with timesteps embedding and fed into the diffusion model. Consequently, the inputs for our proposed CookingDiffusion encompass procedural prompts, timesteps, as well as conditional prompts, and noisy latent variables, similar to the setup in traditional text-to-image generation tasks. Specifically, our utilization of pre-trained text and vision encoders based on CLIP enables the Text Memory Net and Image Memory Net to adapt to procedural prompts from different modalities within a unified structure. As is widely acknowledged, CLIP serves as a common bridge between textual and visual modalities \cite{shen2021much}, which we believe helps to eliminate the inherent modality-specific attributes of these two types of data. These Memory Nets enable the diffusion model to generate high-quality procedural images while maintaining consistency within the cooking steps in sequential order. Subsequent subsections will provide a detailed explanation of each Memory Net.
\label{overvie of MN}

\subsection{Text Memory Net (TMN)}
In Fig.~\ref{overview} (a), we introduce the Text Memory Net (TMN) for text-based procedural representations learning to address the first scenario, where only text prompts are used as inputs. These text prompts also serve as conditional prompts in our diffusion model. For text-based procedural prompts, we firstly feed them $[\mathbf{C}_{i}]_{i=1}^{N}$ into a CLIP-based text encoder to produce the text encodings, written as $[\mathbf{c}_{i}]_{i=1}^{N}$. A simple masked self-attention block is deployed to compute the memory of the encoding sequence $[\mathbf{c}_{i}]_{i=1}^{N}$. The resulting adaptive text memory is then mapped to the same dimension as the time embedding with a linear layer and added to the time embedding. It is worth noting that the parameters of the linear layer need to be initialized to 0 in order to preserve the original pre-trained knowledge intact. Then, we can formulate the entire process of TMN as follows:
\begin{align}
    &\mathbf{c}_{j} = \varepsilon _{text}(\mathbf{C}_{j}), j=1, \cdots, N, \\ 
    &\mathbf{m}_{j}^{text} = SelfAtten([\mathbf{c}_{1}, \cdots, \mathbf{c}_{j-1}]), j=1, \cdots, N,\\
    &\mathbf{e}_{j}^{text} = \mathbf{t}_{j}+Linear(\mathbf{m}_{j}^{text}), j=1, \cdots, N,
\end{align}
where $\varepsilon_{text}$ refers to the text encoder with an MLP block, $\mathbf{m}_{j}^{text}$ and $\mathbf{t}_{j}$ denote the learned text-based procedural representation, and the time embedding of the $j$-th step respectively, and $\mathbf{e}_{j}^{text}$ is as the final input of the UNet within the Stable Diffusion.

\subsection{Image Memory Net (IMN)}
As depicted in Fig.~\ref{overview} (b), our proposed Image Memory Net (IMN) exhibits a structure similar to that of the TMN. In the second scenario, wherein both visual and text cues are available for procedural image generation, the input for CookingDiffusion includes the text sequence as conditional prompts and the corresponding ground-truth images for learning procedural representations. 
When learning image-based procedural representations, an additional linear operator is required to align the shape of image encoding with the text encoding, in order to further be fused with the TMN to conduct the Multi-Modality Memory Net. The remaining process, similar to learning and involving the text-based procedural representations, can be written as:
\begin{align}
    &\mathbf{i}_{j} = \varepsilon_{img}(\mathbf{I}_{j}^{known}), j=1, \cdots, N,\\ 
    &\mathbf{m}_{j}^{img} = SelfAtten([\mathbf{i}_{1}, \cdots, \mathbf{i}_{j-1}]), j=1, \cdots, N,\\
    &\mathbf{e}_{j}^{img} = \mathbf{t}_{j}+Linear(\mathbf{m}_{j}^{img}), j=1, \cdots, N,
\end{align}
where $\operatorname{\varepsilon}_{img}$ denotes the image encoder with an MLP block for dimension alignment and $\mathbf{m}_{j}^{img}$ is the learned image-based procedural representation in $j$-th step. 

\subsection{Multi-modalilty Memory Net (MMN)}
There is a special scenario for the cooking procedural image generation task where the procedural prompts are multi-modal. In practice, it is common that part of the cooking steps contain corresponding images. Notably, these steps with corresponding images available are valuable for providing essential details such as background and objects for our procedural generative task. Thus, both textual and visual cues should be considered during the learning of multi-modal procedural representations in this scenario. TMN and IMN have demonstrated their ability to successfully generate high-quality and consistent procedural images using prompts from different modalities. This capability allows them to address this particular scenario, and we combine them to form the Multi-modality Memory Net (MMN).

In the initial step, the textual descriptions and images of the steps are encoded using the CLIP-based text encoder and vision encoder, respectively. Subsequently, these encodings are aligned in the same dimension using an MLP block. The encodings from different modalities are then combined based on their positions within the step sequence, serving as the input to the self-attention block. Finally, the resulting multi-modal representation is incorporated into the time embedding through a linear layer initialized to 0. Importantly, since the position of the steps with available images may vary and not always occur at the beginning, the memory operates in a bi-directional manner. In essence, the procedural prompt for each step is derived from the entire encoding sequence. In summary, the MMN is an amalgamation of the previously mentioned TMN and IMN, and its detailed structure is illustrated in Fig.~\ref{overview} (c).

\section{Improved baselines}
\begin{figure*}[!t]
\centering
\includegraphics[width=0.85\linewidth]{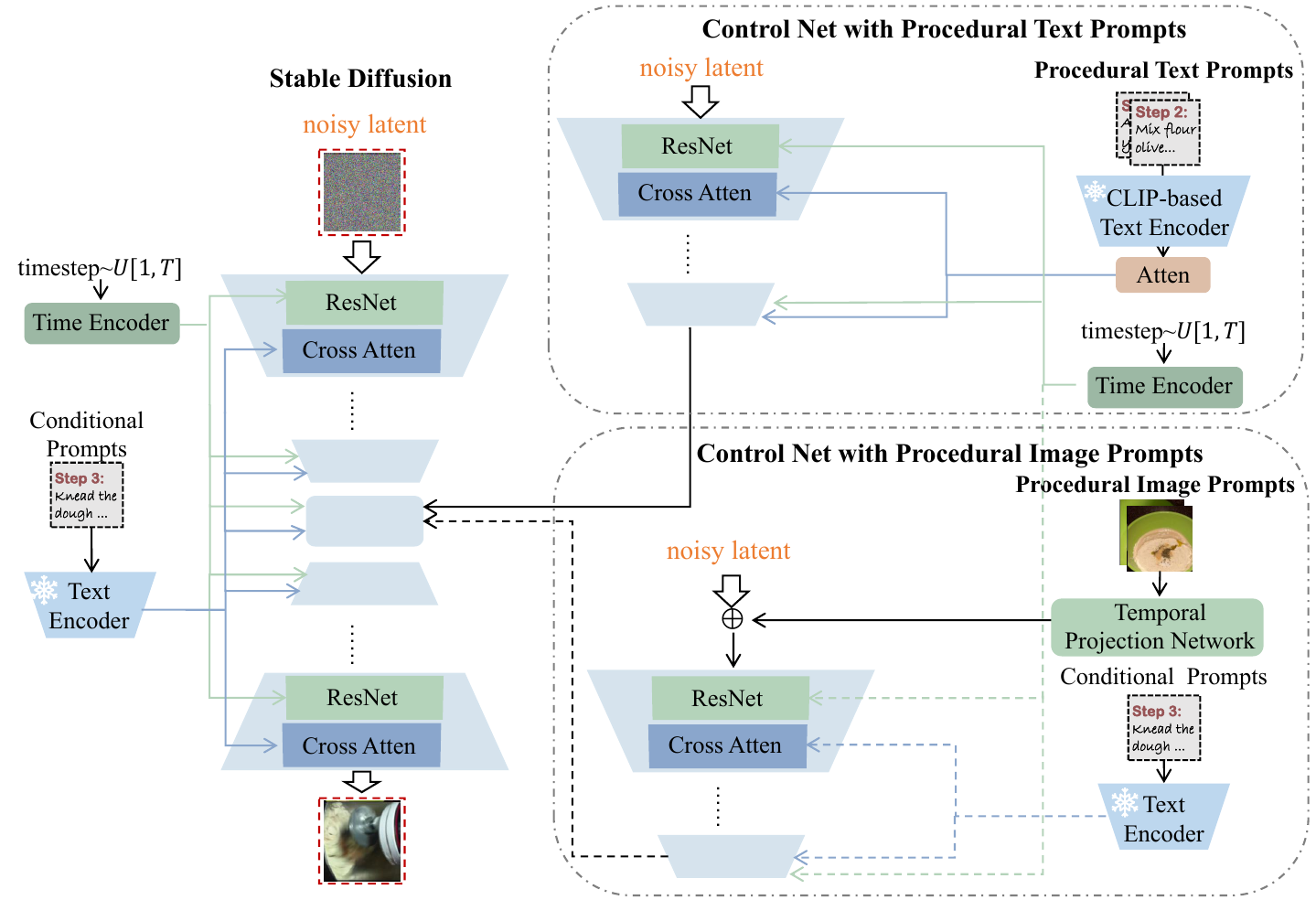}
\caption{Overview of Control Net with procedural text and image prompts. Various modifications are implemented on the Control Net to facilitate the learning and involvement of procedural prompts.}
\label{control_net model}
\end{figure*}

 To further verify the effectiveness of the proposed CookingDiffusion, we also adapt other generative models, such as StackGAN~\cite{zhang2017stackgan}, VQ Diffusion~\cite{Gu_2022_CVPR} and Control Net~\cite{zhang2023adding} for the procedural image generation task as improved baselines for comparison.~In the StackGAN-based method,~procedural representation is obtained through an auxiliary classification task, fed into StackGAN for image generation. In the VQ Diffusion-based method, we apply the proposed TMN and IMN. In the Control Net-based method, we modify its DownBlock copy block and projection network to learn text-based and image-based procedural representations, respectively. The details of these methods are listed below.
\subsection{StackGAN-based Method}
\label{discuss}
Given the challenging nature of training GANs, we introduce an auxiliary task to acquire procedural representations for StackGAN generation, rather than processing procedural prompts in an end-to-end manner. Leveraging the label for each recipe in the YouCookII dataset, we employ a classification task as the auxiliary task. In the StackGAN-based method, text-based and image-based procedural representations are obtained through a GRU module. This module calculates the memory of text encoding and image encoding derived from the CLIP-based text encoder and vision encoder. The final classification result is generated by a classification head on the last hidden state of the GRU. Then, the procedural representation learned from this auxiliary task seamlessly integrates with the encoding of conditional prompts to guide the image generation process.

\subsection{VQ Diffusion-based Method}
The VQ Diffusion closely resembles the structure of Stable Diffusion, initially, it encodes images, compressing them into a low-dimensional discrete space through the VQ VAE. The subsequent diffusion and reverse processes operate within this latent space. VQ Diffusion also includes a text encoder and a diffusion image encoder. The text encoder encodes conditional prompts, while the diffusion image encoder utilizes this conditional encoding, noisy latent variables, and the current timestep to predict the initial latent variable. The cross-attention in the diffusion image encoder incorporates conditional encoding, and the AdaLN layer injects the current timestep into the network. Consequently, we integrate our proposed TMN and IMN directly into VQ Diffusion to process procedural text and image prompts. Similar to the proposed CookingDiffusion, our Memory Net serves as a parallel module with the time encoder. A zero-initialized linear layer combines the time embedding and learned procedural representations, ultimately feeding into the AdaLN layer to guide VQ Diffusion in generating consistent images.

\subsection{Control Net-based Method}
As shown in Fig.~\ref{control_net model}, in the Control Net-based method, we adapt the projection network and the Downblock copy block to incorporate procedural prompts into the generation process, ensuring the consistency of the generated images. The specific modifications for procedural text and image prompts are outlined below.
\label{memory_net method}
\subsubsection{Control Net with procedural text prompts}
Given the text sequence $[\mathbf{C}_{j}]_{j=1}^{i-1}$ as procedural prompts of the step $i$, we also employ the CLIP-based text encoder and the attention block to derive the procedural representation. However, as these text-based procedural representations lack a two-dimensional structure, directly inputting them into the projection network of Control Net is unfeasible. Consequently, we substitute conditional prompts with the acquired text-based procedural representations as one of the inputs for the Downblock copy block, rendering the projection network unnecessary. Additionally, the noisy latent variables fed into Stable Diffusion are also directed into the Downblock copy block. Finally, the outputs of the Downblock copy block are fed to the Middleblock to guide the generation process.

The training strategy, when equipped with procedural prompts, is similar to the Control Net. The Downblock copy is initialized with parameters from the pre-trained Stable Diffusion to leverage its robust generative capability. Furthermore, during training, only the parameters of the attention block in the Downblock copy are updated to mitigate computational costs.

\subsubsection{Control Net with procedural image prompts}
\label{control_net method}
\begin{figure}[!t]
\centering
\includegraphics[width=0.8\linewidth]{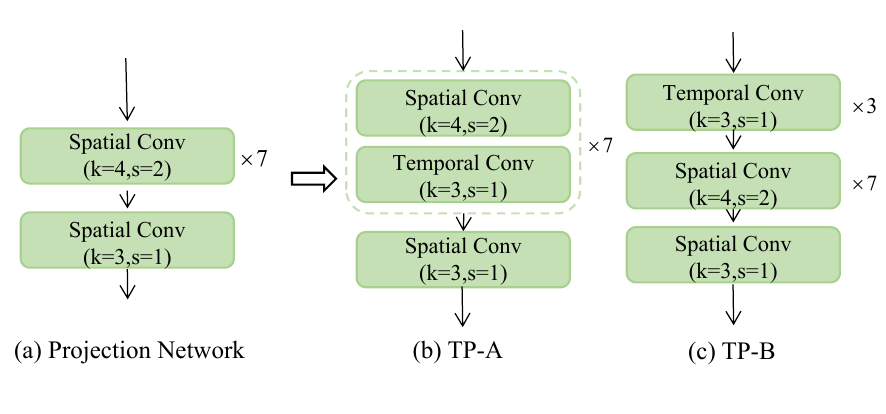}
\caption{(a) is the architecture of the original projection network, and (b), (c) are two proposed temporal projection networks. (b). Temporal Projection Network A (TP-A) (c). Temporal Projection Network B (TP-B). For the sake of simplicity, the activation functions have been omitted here.}
\label{control_net_convt}
\Description{The architecture of the original projection network and the TP network.}
\end{figure}

Illustrated in Fig.~\ref{control_net model}, when processing procedural image prompts, we adapt the projection network to learn image-based procedural representations. The original projection network $P(\cdot)$ in the Control Net, consisting of 7 spatial convolutions and activation functions, serves as an image encoder for feature extraction. We introduced masked temporal convolutions into $P(\cdot)$ to compute image-based procedural representations. The resulting projection network with temporal convolutions is denoted as the temporal projection network, i.e., $TP(\cdot)$. In Fig.~\ref{control_net_convt}, we propose two distinct variations of the temporal projection networks: (1) adding temporal convolutions after each spatial convolution within the original projection network, denoted as $TP-A(\cdot)$, and (2) placing the stacked temporal convolutions and activations before the original projection network, denoted as $TP-B(\cdot)$. 

To encode the correlation between the noisy latent variable and procedural representations, capturing consistency among steps to guide the subsequent decoding process, we input the summation of noisy latent variables and learned image-based procedural representations into the Downblock copy block. During training, consistent with the Control Net's strategy, the Downblock copy is initialized with the parameters of the pre-trained Stable Diffusion and we freeze the Stable Diffusion and only update the parameters of the $TP(\cdot)$ and Downblock copy block.

\section{Experiment}
To validate the effectiveness of our CookingDiffusion with the proposed Memory Nets, we conduct experiments based on pre-processed YouCookII data for cooking procedural image generation. We measure the quality of the generated images through two metrics: Fréchet Inception Distance(FID) \cite{heusel2017gans} to evaluate the authenticity of the generated images and a newly proposed metric named Average Procedure Consistency to assess the consistency among images within a procedure. Additionally, we provide visualizations of some generated cases to showcase the excellent performance of our CookingDiffusion.~Throughout the experiments, we mainly evaluate the model's ability to learn task-specific procedural representations and incorporate them into the procedural image generation process.

\subsection{Pre-processing on YouCookII Dataset}
The YouCookII dataset is a valuable resource for our research and contains thousands of cooking videos from YouTube. Each video is enriched with detailed step-by-step instructions and timestamps, marking the beginning and end of each step. In our experiments, we omitted videos that were no longer available on YouTube or could not be segmented into frame sequences, resulting in a dataset comprising 1185 training videos (9171 text-image pairs) and 446 validation videos (3420 text-image pairs).

To cater to the needs of cooking procedural image generation, these videos undergo preprocessing into sequences of procedural text-image pairs. This involves selecting keyframes that best match each step, with a pre-trained CLIP model assisting in frame selection based on similarity score. These keyframes often capture both actions and their partial outcomes, such as a frame for ``slicing tomatoes'' showing not only a hand with a knife but also partially sliced tomatoes. We resize all images to a resolution of $256\times256$. Due to a perceived lack of samples, we evaluated CookingDiffusion on both the validation and training sets, following previous practices \cite{yu2023video, skorokhodov2022stylegan}.


\subsection{Metrics}
\subsubsection{Fréchet Inception Distance (FID)}
FID \cite{heusel2017gans} is a common metric to assess the quality of generated images. It quantifies the distance between the real images and generative images by deploying Inception-v3 \cite{szegedy2016rethinking}. A lower FID value indicates that the distribution of generated images is closer to that of the real images, thereby representing a higher capability of the generative model. 
\subsubsection{Average Procedure Consistency (Avg-PCon)}
Although FID is effective in quantifying the model's capability to generate authentic and diverse images, it lacks the ability to measure the consistency among the generated step images within a procedure, which is a significant aspect of our task. As the procedural images are from different steps in one recipe, we do not require continuity between adjacent frames as video generation \cite{yin2023nuwa,yu2023video,villegas2022phenaki,ho2022video,ho2022imagen}. In this case, we propose Average Procedure Consistency (Avg-PCon) to evaluate whether the consistency of generated images and their step texts is aligned.

Avg-PCon is based on the CLIP scores between the cooking steps and the generated images. For the text descriptions $[\mathbf{C}_{1}, \cdots, \mathbf{C}_{N}]$ in a recipe and corresponding generated step images $[\mathbf{I}_{1}^{gen}, \cdots, \mathbf{I}_{N}^{gen}]$, we calculate the CLIP scores between each image and textual description of other steps. For instance, given $\mathbf{I}_{i}^{gen}$, we compute the CLIP score between $\mathbf{I}_{i}^{gen}$ and $\mathbf{C}_{j} (j=1,\cdots,i-1,i+1,\cdots, N)$, denoted as $\Braket{\mathbf{I}_{i}^{gen}, \mathbf{C}_{j}}$, to measure the consistency between current step image with all the other step texts. It is worth noting that we do not solely pursue the high value of $\Braket{\mathbf{I}_{i}^{gen}, \mathbf{C}_{j}}$. If the similarity between $\mathbf{C}_{i}$ and $\mathbf{C}_{j}$, denoted as $\Braket{\mathbf{C}_{i}, \mathbf{C}_{j}}$, is low, the corresponding CLIP score $\Braket{\mathbf{I}_{i}^{gen}, \mathbf{C}_{j}}$ should also be low. Therefore, we utilize normalized $\Braket{\mathbf{C}_{i}, \mathbf{C}_{j}}$ as the weight of $\Braket{\mathbf{I}_{i}^{gen}, \mathbf{C}_{j}}$ to evaluate the consistency between $\mathbf{I}_{i}^{gen}$ and the entire procedure. We denote the procedure consistency for $\mathbf{I}_{i}^{gen}$ as $P_{i}$, and it can be formalized as:
\begin{align}
    &P_{i} = \sum_{j=1,j\neq i}^{N} \Braket{\mathbf{C}_i, \mathbf{C}_{j}}^{norm}\cdot\Braket{\mathbf{I}_{i}^{gen}, \mathbf{C}_j}, \\
    &P = Avg(\left [P_{1}, \cdots, P_{N} \right ]),
\end{align}
where $\left\langle \mathbf{C}_{i}, \mathbf{C}_{j}\right\rangle^{norm}$ is the normalized text similarity, and $P$ denotes the consistency of the procedure, named \textbf{Procedure Consistency}. We finally compute the average of $P$ for each procedure as \textbf{Average Procedure Consistency} of the entire dataset. According to this metric, a higher value indicates better consistency among the generated step images.

\subsection{Performance comparison in scenarios 1 and 2}
\label{exp_s1_s2}

\begin{table}[t]
\centering
\small
\caption{Performance comparison of different methods on the procedural image generation task in Scenarios 1 and 2. ``$\uparrow$" indicates higher is better and ``$\downarrow$" indicates lower is better. Since the ``baseline'' methods cannot process procedural prompts, they only use the text description at the current step as input.}
\label{TMN&IMN result}
\resizebox{0.95\columnwidth}{!}{
\begin{tabular}{cl|cc|cc}
\toprule[1pt]
\multicolumn{2}{c|}{}                                                                                   & \multicolumn{2}{c|}{Training Set}                                                 & \multicolumn{2}{c}{Val Set}                                                       \\ \cline{3-6} 
\multicolumn{2}{c|}{\multirow{-2}{*}{Method}}                                                           & FID    $\downarrow$                                 & Avg-PCon  $\uparrow$                             & FID      $\downarrow$                               & Avg-PCon     $\uparrow$                           \\ \midrule[1pt]
\multicolumn{1}{c|}{}                             & StackGAN~\cite{zhang2017stackgan}                                          & 92.748                                  & 13.251                                  & 181.940                                 & 13.327                                  \\
\multicolumn{1}{c|}{}                             & \cellcolor{green!20!yellow!5!}VQ Diffusion~\cite{Gu_2022_CVPR}                 & \cellcolor{green!20!yellow!5!}67.588          & \cellcolor{green!20!yellow!5!}16.987          & \cellcolor{green!20!yellow!5!}76.461          & \cellcolor{green!20!yellow!5!}16.972          \\
\multicolumn{1}{c|}{}                             & \cellcolor{green!20!yellow!5!}Pretrained Stable Diffusion~\cite{rombach2022high}  & \cellcolor{green!20!yellow!5!}51.040          & \cellcolor{green!20!yellow!5!}17.791          & \cellcolor{green!20!yellow!5!}62.588          & \cellcolor{green!20!yellow!5!}17.899          \\
\multicolumn{1}{c|}{\multirow{-4}{*}{Baseline}}   & \cellcolor{green!20!yellow!5!}Fine-tuned Stable Diffusion~\cite{rombach2022high}  & \cellcolor{green!20!yellow!5!}29.585          & \cellcolor{green!20!yellow!5!}18.474          & \cellcolor{green!20!yellow!5!}40.394          & \cellcolor{green!20!yellow!5!}18.470           \\ \hline
\multicolumn{1}{c|}{}                             & StackGAN with procedural text prompts                                       & 73.151                                  & 17.005                                  & 84.137                                  & 16.823                                  \\
\multicolumn{1}{c|}{}                             & \cellcolor{green!20!yellow!5!}VQ Diffusion with TMN       & \cellcolor{green!20!yellow!5!}62.763          & \cellcolor{green!20!yellow!5!}17.463          & \cellcolor{green!20!yellow!5!}70.922          & \cellcolor{green!20!yellow!5!}17.593          \\
\multicolumn{1}{c|}{}                             & \cellcolor{green!20!yellow!5!}Control Net with procedural text prompts                  & \cellcolor{green!20!yellow!5!}36.112       & \cellcolor{green!20!yellow!5!}16.485          & \cellcolor{green!20!yellow!5!}46.739          & \cellcolor{green!20!yellow!5!}16.600           \\
\multicolumn{1}{c|}{\multirow{-4}{*}{Scenario 1}} & \cellcolor{green!20!yellow!5!}CookingDiffusion with TMN   & \cellcolor{green!20!yellow!5!}24.336          & \cellcolor{green!20!yellow!5!}18.648          & \cellcolor{green!20!yellow!5!}34.547          & \cellcolor{green!20!yellow!5!}18.541          \\ \hline
\multicolumn{1}{c|}{}                             & StackGAN with procedural image prompts                                      & 205.138                                 & 14.995                                  & 211.266                                 & 15.051                                  \\
\multicolumn{1}{c|}{}                             & \cellcolor{green!20!yellow!5!}VQ Diffusion with IMN       & \cellcolor{green!20!yellow!5!}69.914          & \cellcolor{green!20!yellow!5!}17.228          & \cellcolor{green!20!yellow!5!}81.590           & \cellcolor{green!20!yellow!5!}16.982          \\
\multicolumn{1}{c|}{}                             & \cellcolor{green!20!yellow!5!}Control Net with procedural image prompts                 & \cellcolor{green!20!yellow!5!}27.468          & \cellcolor{green!20!yellow!5!}17.385          & \cellcolor{green!20!yellow!5!}38.786          & \cellcolor{green!20!yellow!5!}17.765          \\
\multicolumn{1}{c|}{\multirow{-4}{*}{Scenario2}}  & \cellcolor{green!20!yellow!5!}\textbf{CookingDiffusion with IMN}   & \cellcolor{green!20!yellow!5!}\textbf{23.156} & \cellcolor{green!20!yellow!5!}\textbf{18.707} & \cellcolor{green!20!yellow!5!}\textbf{34.294} & \cellcolor{green!20!yellow!5!}\textbf{18.755} \\ \bottomrule[1pt]
\end{tabular}}
\end{table}
\begin{figure*}[!t]
\includegraphics[width=\linewidth]{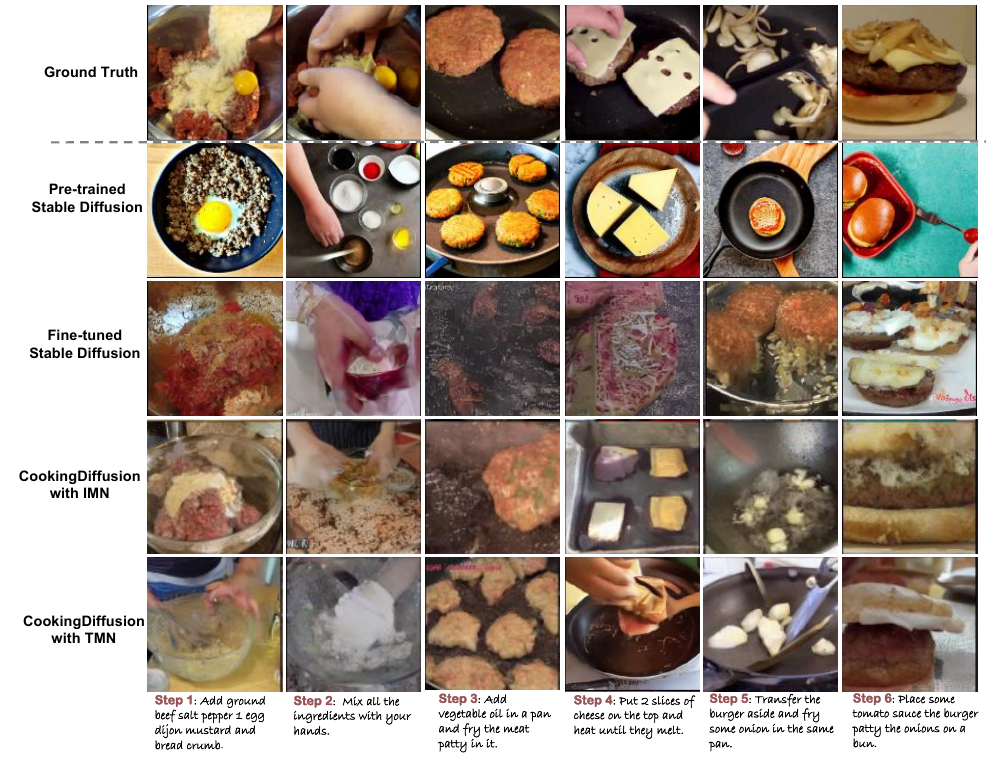}
\centering
\caption{Comparison of the generated procedural images using CookingDiffusion with TMN and IMN and Stable Diffusion.}
\label{modality memory net}
\Description{Experiments result of the CookingDiffusion with the UMN.}
\end{figure*}

In this section, we conduct a comprehensive performance analysis of diverse methods in scenarios 1 and 2. The training protocol for CookingDiffusion with Text Memory Net and Image Memory Net adheres to the same hyperparameter settings. Specifically, we fix the number of diffusion process timesteps at 1000, while the learning rate for the denoising model is set to 1e-5. A total of 75 epochs are conducted for training, utilizing Stable Diffusion V1.5 as the base model for CookingDiffusion.  

To assess the efficacy of procedural prompts for different generative backbones, we directly train StackGAN and VQ Diffusion on the YouCookII dataset as the baseline. It's essential to highlight that the training of the Control Net requires additional control information for guiding the generation process. Given the inherent challenge of directly training the Control Net on the dataset, and considering that the parameters of the Stable Diffusion component remain frozen during the training process of the Control Net, we use the pre-trained Stable Diffusion as the baseline to assess the performance of the Control Net under various procedural prompts. Additionally, for CookingDiffusion, we performed fine-tuning on the Stable Diffusion using the YouCookII dataset as the baseline.


\begin{table*}[!t]
\caption{Results of Multi-modality Memory Net under different percentages $p$ of available images in distinct positions. The term ``ordered-available'' refers to available images in the initial steps, while ``random-available'' indicates that the position of available images is random.} \label{mm result}
\centering
\resizebox{0.95\columnwidth}{!}{
\begin{tabular}{c|c|cccc|cccc}
\toprule[1pt]
\multirow{3}{*}{\textbf{$p$ for training}} & \multirow{3}{*}{\textbf{$p$ for validation}} & \multicolumn{4}{c|}{\textbf{ordered-available}}                                            & \multicolumn{4}{c}{\textbf{random-available}}                                             \\
                                         &                                            & \multicolumn{2}{c}{\textbf{training set}} & \multicolumn{2}{c|}{\textbf{validation set}} & \multicolumn{2}{c}{\textbf{training set}} & \multicolumn{2}{c}{\textbf{validation set}} \\
                                         &                                            & \textbf{FID} $\downarrow$       & \textbf{Avg-PCon} $\uparrow$    & \textbf{FID} $\downarrow$        & \textbf{Avg-PCon} $\uparrow$        & \textbf{FID} $\downarrow$       & \textbf{Avg-PCon} $\uparrow$       & \textbf{FID} $\downarrow$       & \textbf{Avg-PCon} $\uparrow$        \\ \midrule[1pt]
0.2                                      & 0.2                                        & 25.894             & 18.750               & 38.000               & 18.938                & 24.607             & 18.587               & \textbf{35.091}     & 18.920                \\
0.2                                      & random                                     & 26.783             & 18.822               & 40.984               & 18.845                & 24.772             & 18.678               & 36.702              & \textbf{18.922}       \\
\hline
0.3                                      & 0.3                                        & \textbf{24.449}    & \textbf{19.200}      & 37.803               & \textbf{19.518}       & 24.014             & \textbf{18.610}      & 39.165              & 18.591                \\
0.3                                      & random                                     & 25.410             & 18.610               & \textbf{36.777}      & 19.362                & \textbf{23.766}    & 18.590               & 37.940              & 18.544                \\
\hline
0.4                                      & 0.4                                        & 26.116             & 18.905               & 41.471               & 18.838                & 31.234             & 18.502               & 44.686              & 18.708                \\
0.4                                      & random                                     & 24.890             & 18.665               & 37.752               & 18.783                & 27.064             & 18.368               & 40.327              & 18.520                \\ \bottomrule[1pt]
\end{tabular}}
\end{table*}

\begin{table*}[!t]
\caption{Results of Multi-modality Memory Net under different percentages $p$ of available images in distinct positions. In this scenario, the steps with available images also have corresponding text instructions, and both of them are used for procedural representation learning.} \label{mm_v2 result}
\centering
\resizebox{0.95\columnwidth}{!}{
\begin{tabular}{c|c|cccc|cccc}
\toprule[1pt]
\multirow{3}{*}{\textbf{$p$ for training}} & \multirow{3}{*}{\textbf{$p$ for validation}} & \multicolumn{4}{c|}{\textbf{ordered-available}}                                            & \multicolumn{4}{c}{\textbf{random-available}}                                             \\
                                         &                                            & \multicolumn{2}{c}{\textbf{training set}} & \multicolumn{2}{c|}{\textbf{validation set}} & \multicolumn{2}{c}{\textbf{training set}} & \multicolumn{2}{c}{\textbf{validation set}} \\
                                         &                                            & \textbf{FID} $\downarrow$       & \textbf{Avg-PCon} $\uparrow$    & \textbf{FID} $\downarrow$        & \textbf{Avg-PCon} $\uparrow$        & \textbf{FID} $\downarrow$       & \textbf{Avg-PCon} $\uparrow$       & \textbf{FID} $\downarrow$       & \textbf{Avg-PCon} $\uparrow$        \\ \midrule[1pt]
0.2                                      & 0.2                                        & 24.261             & 18.942               & \textbf{36.499}             & 18.890                  & 23.900             & 18.783               & \textbf{36.885}     & 18.791                \\
0.2                                      & random                                     & 24.894             & 18.950               & 37.913             & 19.105                  & 25.037             & 18.807               & 37.769              & 18.760                \\ \hline
0.3                                      & 0.3                                        & 24.329             & \textbf{19.136}      & 38.627             & \textbf{19.166}         & 24.689             & \textbf{18.944}      & 38.934              & \textbf{18.988}       \\
0.3                                      & random                                     & 29.901             & 18.813               & 43.102             & 18.803                  & 29.990             & 18.610               & 43.766              & 18.627                \\ \hline
0.4                                      & 0.4                                        & 25.577             & 18.829               & 41.240             & 18.894                  & 25.457             & 18.775               & 41.243              & 18.716                \\
0.4                                      & random                                     & \textbf{23.700}    & 18.858               & 37.264             & 18.840                  & \textbf{23.383}    & 18.688               & 37.631              & 18.633                \\  \bottomrule[1pt]
\end{tabular}}
\end{table*}  

Table~\ref{TMN&IMN result}~presents a performance comparison of the aforementioned methods. The StackGAN-based and VQ Diffusion-based methods, as seen in the table, enhance consistency in generated procedural images but struggle to maintain image quality in scenario 2. Notably, the decline in image quality when employing procedural image prompts with StackGAN can be attributed to suboptimal image-based procedural representation quality, as supported by its performance in the auxiliary classification task mentioned in section~\ref{discuss}. The image quality drop observed with VQ Diffusion and procedural image prompts is linked to the difficulty in striking a balance between procedural consistency and image quality. Conversely, the Control Net-based method produces more vibrant images but falls short in ensuring consistency within a procedure. In contrast, our proposed CookingDiffusion outperforms all the mentioned methods in both FID and Avg-PCon across scenarios 1 and 2. This suggests that CookingDiffusion achieves the optimal trade-off between procedural consistency and image quality. This result highlights CookingDiffusion's capability to achieve outstanding performance with procedural prompts from different modalities through the use of our proposed simple yet highly effective Memory Nets. Given CookingDiffusion's excellent performance in scenarios 1 and 2, we focus on utilizing it to address scenario 3. 

In addition to evaluating the metrics of generated images, We showcase a comparison between the procedural images generated by our model and those generated by fine-tuned Stable Diffusion, which further provides a visual demonstration of consistency among the procedural images achieved by CookingDiffusion. The visual comparison of the generated examples between our CookingDiffusion and Stable Diffusion is presented in Fig.~\ref{modality memory net}. The comparison clearly demonstrates that CookingDiffusion excels in generating consistently high-quality procedural images. For instance, CookingDiffusion effectively depicts the ingredients mentioned in both step 1 and step 4, while fine-tuned Stable Diffusion falls short in this aspect. Moreover, our CookingDiffusion consistently produces more authentic images, particularly evident from step 3 to step 5. In summary, our CookingDiffusion stands out in its ability to generate exceptional procedural images.

\begin{figure*}[!t]
\includegraphics[width=\linewidth]{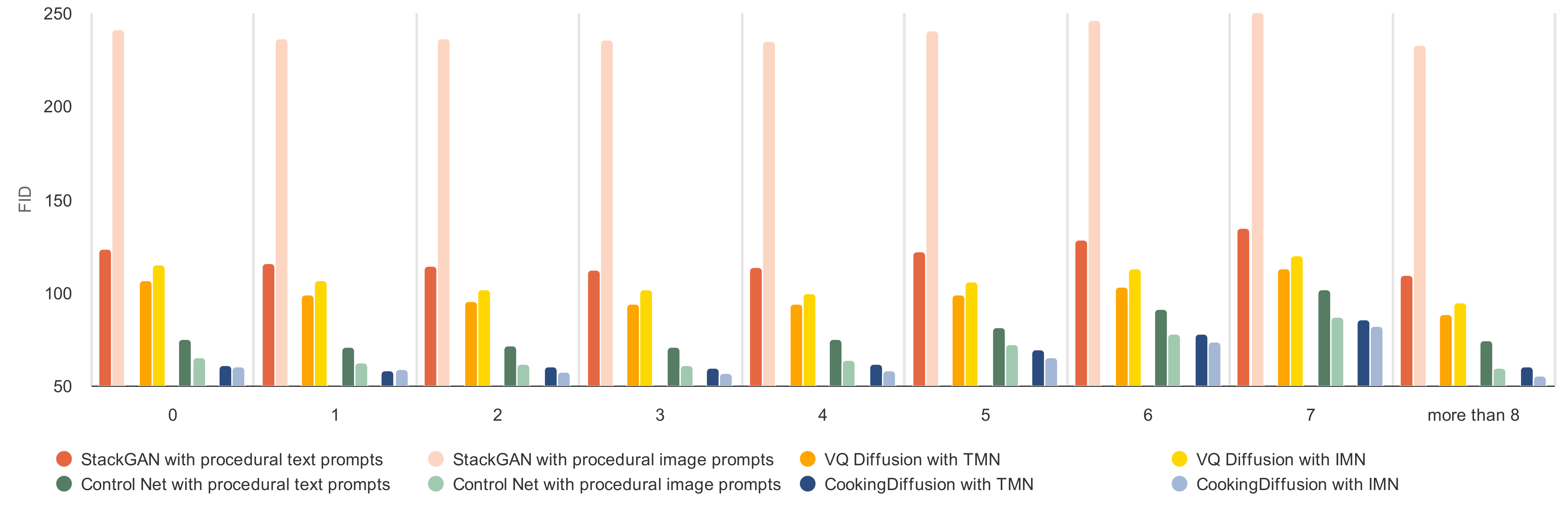}
\centering
\caption{Comparison of FID of the generated procedural images in different positions, i.e., with varying lengths of procedural prompts. The horizontal axis represents the length of procedural prompts.}
\label{fid}
\Description{Comparison of FID of the generated procedural images.}
\end{figure*}

\begin{figure*}[!t]
\includegraphics[width=\linewidth]{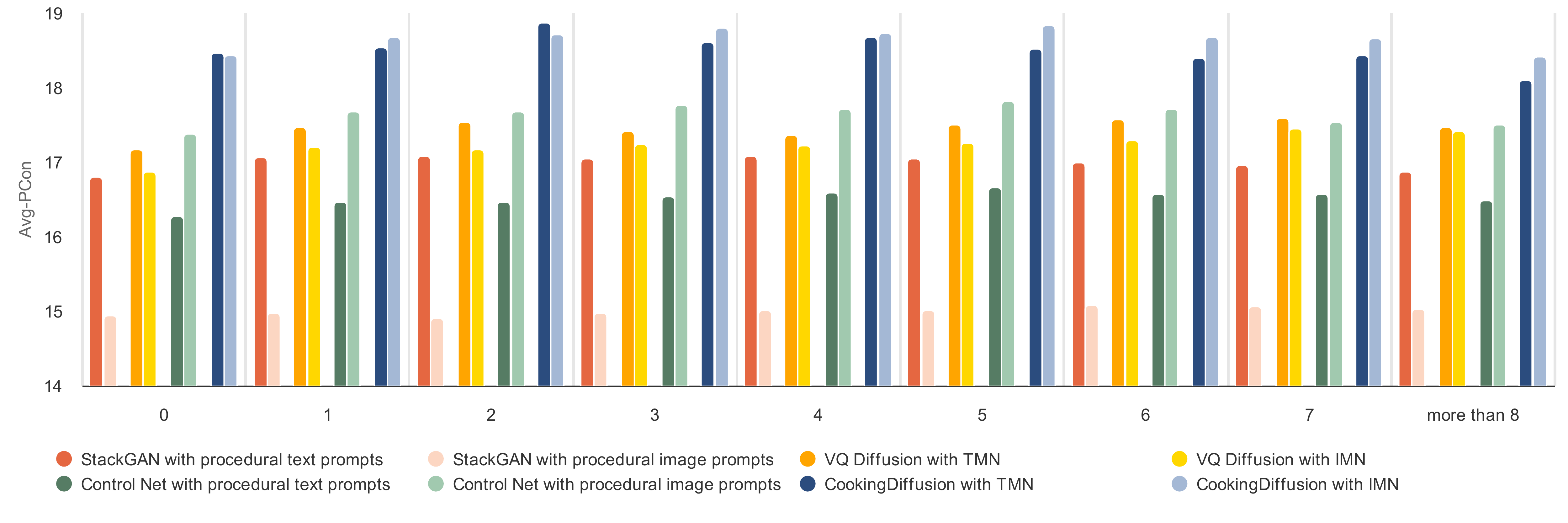}
\centering
\caption{Comparison of Avg-PCon of the generated procedural images in different positions.}
\label{avg-pcon}
\Description{Comparison of Avg-PCon of the generated procedural images.}
\end{figure*}
In addition to assessing the FID and Avg-PCON metrics for all generated images, demonstrating the proficiency of CookingDiffusion with TMN and IMN in producing high-quality and consistent procedural images in scenarios 1 and 2, we further explore the metrics concerning images in distinct positions, i.e., images generated with varying lengths of procedural prompts. We categorize the generated images from different methods based on their historical step lengths and evaluate the metrics respectively, as illustrated in Fig.~\ref{fid} and Fig.~\ref{avg-pcon}. Due to the insufficient quantity of recipes with more than 8 steps, making it impractical to calculate FID for images with more than 8 historical steps respectively, we assess the overall FID and Avg-PCon for all such generated images labeled as ``more than 8" in Fig.~\ref{fid} and Fig.~\ref{avg-pcon}. 

The results clearly demonstrate the remarkable performance of our proposed CookingDiffusion with TMN and IMN in generating step images across different positions, as indicated by both FID and Avg-PCon. This underscores the effectiveness of TMN and IMN in processing procedural prompts of various lengths, thereby further confirming the superiority of our CookingDiffusion with TMN and IMN in the procedural image generation task. 

\subsection{Performance of MMN in scenario 3}
\label{MMN}
During the training process of the Multi-modality Memory Net, an additional step is performed on the YouCookII dataset to acquire multi-modal procedural prompts. To simulate the multi-modal scenario where a certain percentage of steps have available images, we discard a certain percentage of the step texts and replace them with the corresponding ground-truth step images. We simulate two common situations: ``ordered-available'', where the available images are in the initial steps, and ``random-available'', where the available images are randomly distributed throughout the recipe. 
During the evaluation process, we follow the same approach to obtain multi-modal sequences. Table~\ref{mm result} lists the results using multi-modal procedural prompts with different values of $p$ for training, specifically 0.2, 0.3, and 0.4, to examine the influence of the percentage of visual cues in the input sequences on the model's performance. During the evaluation, we have two different settings. The first one sets the same value of $p$ as training for evaluation. Considering the percentage of available images within a recipe can vary in practice, the second one randomly selects $p$ from the interval $(0, 0.5]$ for each recipe during evaluation.

The results presented in Table~\ref{mm result} indicate that our CookingDiffusion with MMN is robust for both evaluation settings.~Furthermore, in both ``ordered-available'' and ``random-available'' cases, increasing $p$ from 0.2 to 0.3 only leads to a slight change in the model's performance. However, when $p$ is further increased to 0.4, there is a noticeable decrease in the quality of the generated images, especially on the validation set. We attribute this decrease in performance to discarding too many texts, which reduces the number of text-image pairs available for training.

 Additionally, as steps with available images may also contain text instructions, we also investigate utilizing both text and images of these steps. Similarly, $p$ denotes the percentage of available images; however, we retain the text instructions for these steps to simulate the aforementioned case, resulting in the procedural prompts sequence denoted as $[\textbf{C}_{i}, \textbf{I}_{m_{j}}^{known}]; i=1, \cdots, N, j=1, \cdots, N_{2}$, where $i$ and $m_{j}$ represent the position of textual and available image steps, respectively. Token mixing is applied to the text and image encodings at position $m_{j}$ using an MLP block to integrate the visual and textual context of these steps, denoted as $\textbf{f}_{m_{j}}$. Consequently, the input of our self-attention block of our MMN in this scenario is $[\textbf{c}_{n_{i}}, \textbf{f}_{m_{j}}]; i=1,\cdots, N_{1}, j=1, \cdots, N_{2}$, where $n_{i}$ refers to the position without available images.

Similarly, we investigate two situations involving ``ordered-available" and ``random-available", assessing the impact of varying percentages of available images on the results. Table~\ref{mm_v2 result} displays the outcomes. Much like the performance illustrated in Section~\ref{MMN}, our MMN also exhibits robustness when trained on the dataset with different percentages of available images in this scenario. It is evident that leveraging both available images and corresponding text instructions for procedural representation learning yields overall better performance, particularly noticeable as $p$ increases to 0.4, when text-image pairs may be insufficient.

\subsection{Content Manipulation}
\begin{figure}[!t]
\includegraphics[width=0.8\linewidth]{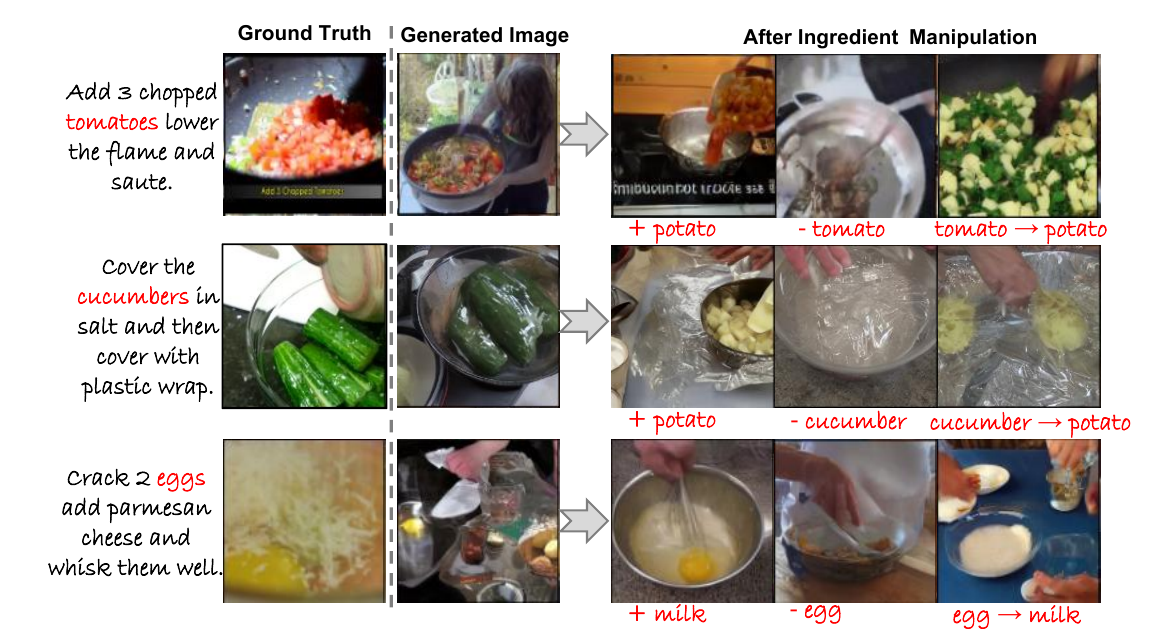}
\centering
\caption{Ingredient manipulation results by adding new ingredients, removing existing ingredients or replacing the ingredients.}
\label{ingredient modify}
\Description{Ingredient manipulation results.}
\end{figure}

\begin{figure}[t]
\includegraphics[width=0.9\linewidth]{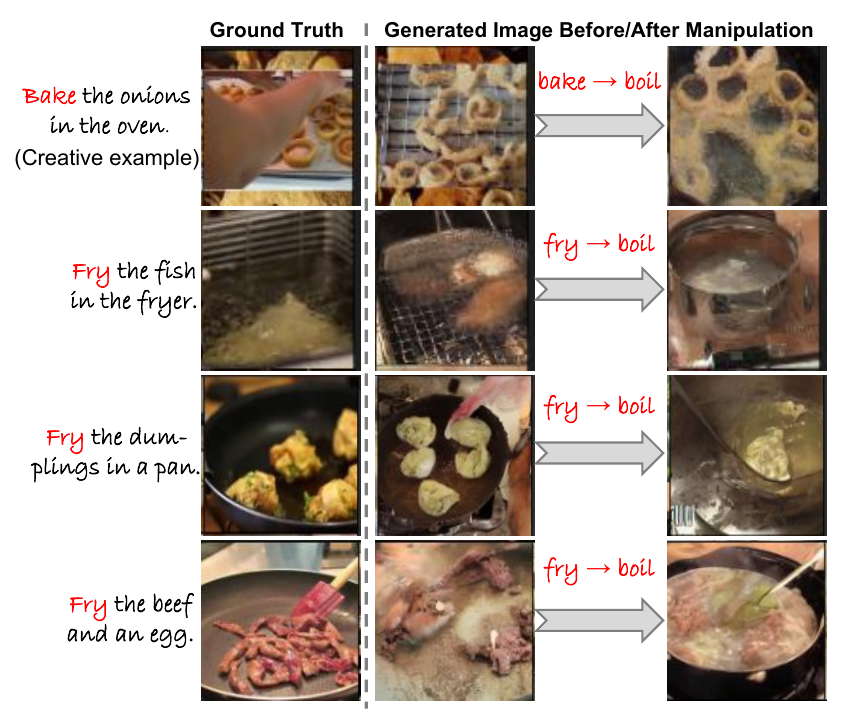}
\caption{Manipulation results by replacing cooking methods, e.g., replacing ``bake'' and ``fry'' with ``boil''.}
\label{cooking method modify}
\Description{Cooking method manipulation.}
\end{figure}

In this section, we examine the capability of CookingDiffusion to manipulate ingredients and cooking methods within the steps of a recipe. On the one hand, ingredients in certain steps could be manipulated, which includes adding a new ingredient, removing the existing ingredient, or even replacing the ingredient. As shown in Fig.~\ref{ingredient modify}, we have successfully demonstrated the versatility of our manipulation technique with various ingredients such as tomatoes and eggs. When any of these ingredients are individually removed, the resulting generated images accurately reflect their absence. Moreover, our method allows for the replacement of one ingredient with another, as exemplified by modifying from tomatoes to potatoes, and eggs to milk. Furthermore, we extend our exploration to include combinations of ingredients. Introducing potatoes to tomatoes or milk to eggs yields images that combine both the original and added ingredients effectively. In a similar manner, we investigate the manipulation of cooking methods, as presented in Fig.~\ref{cooking method modify}. To maintain grammatical correctness and preserve semantic coherence, we focus solely on replacing existing cooking methods with new ones. Though it is important to acknowledge that certain replacements, such as changing ``bake'' to ``boil'', might not correspond to real-world scenarios. Nevertheless, our CookingDiffusion model demonstrates its capability to handle such creative examples and produce desired manipulation results successfully.

\section{Conclusion}
In this paper, we have introduced a distinctive task known as ``cooking procedural image generation'', which significantly diverges from the conventional text-to-image generation task.~It not only strives to generate photo-realistic images corresponding with the cooking steps, but also insists on maintaining consistency within these steps in sequential order.~We have designed CookingDiffusion, a model tailored specifically for cooking procedural image generation.~The core of CookingDiffusion lies in three distinct types of Memory Nets, which facilitate the learning and involvement of procedural prompts in different scenarios.~Our model has been extensively evaluated on the pre-processed YouCookII dataset, and the experimental results illustrate its efficacy in generating procedural images with exceptional quality and consistency.~Moreover,~CookingDiffusion demonstrated its versatility by performing well with end-to-end learned procedural representations.~These results highlight its significant contributions to cooking procedural image generation and pave the way for further advancements in the food domain.~While encouraging,~CookingDiffusion only considers the procedural generation at the image level without temporal modeling.~In~the~future,~we plan to extend this work to step-wise cooking video~generation.

 


\bibliographystyle{ACM-Reference-Format}
\bibliography{sample-base}


\begin{thebibliography}{62}


\ifx \showCODEN    \undefined \def \showCODEN     #1{\unskip}     \fi
\ifx \showDOI      \undefined \def \showDOI       #1{#1}\fi
\ifx \showISBNx    \undefined \def \showISBNx     #1{\unskip}     \fi
\ifx \showISBNxiii \undefined \def \showISBNxiii  #1{\unskip}     \fi
\ifx \showISSN     \undefined \def \showISSN      #1{\unskip}     \fi
\ifx \showLCCN     \undefined \def \showLCCN      #1{\unskip}     \fi
\ifx \shownote     \undefined \def \shownote      #1{#1}          \fi
\ifx \showarticletitle \undefined \def \showarticletitle #1{#1}   \fi
\ifx \showURL      \undefined \def \showURL       {\relax}        \fi
\providecommand\bibfield[2]{#2}
\providecommand\bibinfo[2]{#2}
\providecommand\natexlab[1]{#1}
\providecommand\showeprint[2][]{arXiv:#2}

\bibitem[Chen et~al\mbox{.}(2024)]%
        {10445245}
\bibfield{author}{\bibinfo{person}{Hong Chen}, \bibinfo{person}{Yipeng Zhang}, \bibinfo{person}{Xin Wang}, \bibinfo{person}{Xuguang Duan}, \bibinfo{person}{Yuwei Zhou}, {and} \bibinfo{person}{Wenwu Zhu}.} \bibinfo{year}{2024}\natexlab{}.
\newblock \showarticletitle{DisenDreamer: Subject-Driven Text-to-Image Generation with Sample-aware Disentangled Tuning}.
\newblock \bibinfo{journal}{\emph{IEEE Transactions on Circuits and Systems for Video Technology}} (\bibinfo{year}{2024}), \bibinfo{pages}{1--1}.
\newblock


\bibitem[Chen and Ngo(2016)]%
        {chen2016deep}
\bibfield{author}{\bibinfo{person}{Jingjing Chen} {and} \bibinfo{person}{Chong-Wah Ngo}.} \bibinfo{year}{2016}\natexlab{}.
\newblock \showarticletitle{Deep-based ingredient recognition for cooking recipe retrieval}. In \bibinfo{booktitle}{\emph{Proceedings of the 24th ACM International Conference on Multimedia}}. \bibinfo{pages}{32--41}.
\newblock


\bibitem[Chen et~al\mbox{.}(2020)]%
        {chen2020study}
\bibfield{author}{\bibinfo{person}{Jingjing Chen}, \bibinfo{person}{Bin Zhu}, \bibinfo{person}{Chong-Wah Ngo}, \bibinfo{person}{Tat-Seng Chua}, {and} \bibinfo{person}{Yu-Gang Jiang}.} \bibinfo{year}{2020}\natexlab{}.
\newblock \showarticletitle{A study of multi-task and region-wise deep learning for food ingredient recognition}.
\newblock \bibinfo{journal}{\emph{IEEE Transactions on Image Processing}}  \bibinfo{volume}{30} (\bibinfo{year}{2020}), \bibinfo{pages}{1514--1526}.
\newblock


\bibitem[Cheng et~al\mbox{.}(2022)]%
        {9656731}
\bibfield{author}{\bibinfo{person}{Jun Cheng}, \bibinfo{person}{Fuxiang Wu}, \bibinfo{person}{Yanling Tian}, \bibinfo{person}{Lei Wang}, {and} \bibinfo{person}{Dapeng Tao}.} \bibinfo{year}{2022}\natexlab{}.
\newblock \showarticletitle{RiFeGAN2: Rich Feature Generation for Text-to-Image Synthesis From Constrained Prior Knowledge}.
\newblock \bibinfo{journal}{\emph{IEEE Transactions on Circuits and Systems for Video Technology}} \bibinfo{volume}{32}, \bibinfo{number}{8} (\bibinfo{year}{2022}), \bibinfo{pages}{5187--5200}.
\newblock


\bibitem[Chhikara et~al\mbox{.}(2024)]%
        {chhikara2024fire}
\bibfield{author}{\bibinfo{person}{Prateek Chhikara}, \bibinfo{person}{Dhiraj Chaurasia}, \bibinfo{person}{Yifan Jiang}, \bibinfo{person}{Omkar Masur}, {and} \bibinfo{person}{Filip Ilievski}.} \bibinfo{year}{2024}\natexlab{}.
\newblock \showarticletitle{Fire: Food image to recipe generation}. In \bibinfo{booktitle}{\emph{Proceedings of the IEEE/CVF Winter Conference on Applications of Computer Vision}}. \bibinfo{pages}{8184--8194}.
\newblock


\bibitem[Dhariwal and Nichol(2021)]%
        {dhariwal2021diffusion}
\bibfield{author}{\bibinfo{person}{Prafulla Dhariwal} {and} \bibinfo{person}{Alexander Nichol}.} \bibinfo{year}{2021}\natexlab{}.
\newblock \showarticletitle{Diffusion models beat gans on image synthesis}.
\newblock \bibinfo{journal}{\emph{Advances in Neural Information Processing Systems}}  \bibinfo{volume}{34} (\bibinfo{year}{2021}), \bibinfo{pages}{8780--8794}.
\newblock


\bibitem[Esser et~al\mbox{.}(2021)]%
        {esser2021taming}
\bibfield{author}{\bibinfo{person}{Patrick Esser}, \bibinfo{person}{Robin Rombach}, {and} \bibinfo{person}{Bjorn Ommer}.} \bibinfo{year}{2021}\natexlab{}.
\newblock \showarticletitle{Taming transformers for high-resolution image synthesis}. In \bibinfo{booktitle}{\emph{Proceedings of the IEEE/CVF Conference on Computer Vision and Pattern Recognition}}. \bibinfo{pages}{12873--12883}.
\newblock


\bibitem[Goodfellow(2016)]%
        {goodfellow2016nips}
\bibfield{author}{\bibinfo{person}{Ian Goodfellow}.} \bibinfo{year}{2016}\natexlab{}.
\newblock \showarticletitle{Nips 2016 tutorial: Generative adversarial networks}.
\newblock \bibinfo{journal}{\emph{arXiv preprint arXiv:1701.00160}} (\bibinfo{year}{2016}).
\newblock


\bibitem[Goodfellow et~al\mbox{.}(2014)]%
        {goodfellow2014generative}
\bibfield{author}{\bibinfo{person}{Ian Goodfellow}, \bibinfo{person}{Jean Pouget-Abadie}, \bibinfo{person}{Mehdi Mirza}, \bibinfo{person}{Bing Xu}, \bibinfo{person}{David Warde-Farley}, \bibinfo{person}{Sherjil Ozair}, \bibinfo{person}{Aaron Courville}, {and} \bibinfo{person}{Yoshua Bengio}.} \bibinfo{year}{2014}\natexlab{}.
\newblock \showarticletitle{Generative adversarial nets}.
\newblock \bibinfo{journal}{\emph{Advances in Neural Information Processing Systems}}  \bibinfo{volume}{27} (\bibinfo{year}{2014}).
\newblock


\bibitem[Gu et~al\mbox{.}(2022)]%
        {Gu_2022_CVPR}
\bibfield{author}{\bibinfo{person}{Shuyang Gu}, \bibinfo{person}{Dong Chen}, \bibinfo{person}{Jianmin Bao}, \bibinfo{person}{Fang Wen}, \bibinfo{person}{Bo Zhang}, \bibinfo{person}{Dongdong Chen}, \bibinfo{person}{Lu Yuan}, {and} \bibinfo{person}{Baining Guo}.} \bibinfo{year}{2022}\natexlab{}.
\newblock \showarticletitle{Vector Quantized Diffusion Model for Text-to-Image Synthesis}. In \bibinfo{booktitle}{\emph{Proceedings of the IEEE/CVF Conference on Computer Vision and Pattern Recognition (CVPR)}}. \bibinfo{pages}{10696--10706}.
\newblock


\bibitem[Han et~al\mbox{.}(2019)]%
        {han2019art}
\bibfield{author}{\bibinfo{person}{Fangda Han}, \bibinfo{person}{Ricardo Guerrero}, {and} \bibinfo{person}{Vladimir Pavlovic}.} \bibinfo{year}{2019}\natexlab{}.
\newblock \showarticletitle{The art of food: Meal image synthesis from ingredients}.
\newblock \bibinfo{journal}{\emph{arXiv preprint arXiv:1905.13149}} (\bibinfo{year}{2019}).
\newblock


\bibitem[Heusel et~al\mbox{.}(2017)]%
        {heusel2017gans}
\bibfield{author}{\bibinfo{person}{Martin Heusel}, \bibinfo{person}{Hubert Ramsauer}, \bibinfo{person}{Thomas Unterthiner}, \bibinfo{person}{Bernhard Nessler}, {and} \bibinfo{person}{Sepp Hochreiter}.} \bibinfo{year}{2017}\natexlab{}.
\newblock \showarticletitle{Gans trained by a two time-scale update rule converge to a local nash equilibrium}.
\newblock \bibinfo{journal}{\emph{Advances in Neural Information Processing Systems}}  \bibinfo{volume}{30} (\bibinfo{year}{2017}).
\newblock


\bibitem[Ho et~al\mbox{.}(2022a)]%
        {ho2022imagen}
\bibfield{author}{\bibinfo{person}{Jonathan Ho}, \bibinfo{person}{William Chan}, \bibinfo{person}{Chitwan Saharia}, \bibinfo{person}{Jay Whang}, \bibinfo{person}{Ruiqi Gao}, \bibinfo{person}{Alexey Gritsenko}, \bibinfo{person}{Diederik~P Kingma}, \bibinfo{person}{Ben Poole}, \bibinfo{person}{Mohammad Norouzi}, \bibinfo{person}{David~J Fleet}, {et~al\mbox{.}}} \bibinfo{year}{2022}\natexlab{a}.
\newblock \showarticletitle{Imagen video: High definition video generation with diffusion models}.
\newblock \bibinfo{journal}{\emph{arXiv preprint arXiv:2210.02303}} (\bibinfo{year}{2022}).
\newblock


\bibitem[Ho et~al\mbox{.}(2020)]%
        {ho2020denoising}
\bibfield{author}{\bibinfo{person}{Jonathan Ho}, \bibinfo{person}{Ajay Jain}, {and} \bibinfo{person}{Pieter Abbeel}.} \bibinfo{year}{2020}\natexlab{}.
\newblock \showarticletitle{Denoising diffusion probabilistic models}.
\newblock \bibinfo{journal}{\emph{Advances in Neural Information Processing Systems}}  \bibinfo{volume}{33} (\bibinfo{year}{2020}), \bibinfo{pages}{6840--6851}.
\newblock


\bibitem[Ho et~al\mbox{.}(2022b)]%
        {ho2022video}
\bibfield{author}{\bibinfo{person}{Jonathan Ho}, \bibinfo{person}{Tim Salimans}, \bibinfo{person}{Alexey Gritsenko}, \bibinfo{person}{William Chan}, \bibinfo{person}{Mohammad Norouzi}, {and} \bibinfo{person}{David~J. Fleet}.} \bibinfo{year}{2022}\natexlab{b}.
\newblock \bibinfo{title}{Video Diffusion Models}.
\newblock
\newblock
\showeprint[arxiv]{2204.03458}~[cs.CV]


\bibitem[Horita et~al\mbox{.}(2019)]%
        {horita2019unseen}
\bibfield{author}{\bibinfo{person}{Daichi Horita}, \bibinfo{person}{Wataru Shimoda}, {and} \bibinfo{person}{Keiji Yanai}.} \bibinfo{year}{2019}\natexlab{}.
\newblock \showarticletitle{Unseen food creation by mixing existing food images with conditional stylegan}. In \bibinfo{booktitle}{\emph{Proceedings of the 5th International Workshop on Multimedia Assisted Dietary Management}}. \bibinfo{pages}{19--24}.
\newblock


\bibitem[Ito et~al\mbox{.}(2018)]%
        {ito2018food}
\bibfield{author}{\bibinfo{person}{Yoshifumi Ito}, \bibinfo{person}{Wataru Shimoda}, {and} \bibinfo{person}{Keiji Yanai}.} \bibinfo{year}{2018}\natexlab{}.
\newblock \showarticletitle{Food image generation using a large amount of food images with conditional gan: ramengan and recipegan}. In \bibinfo{booktitle}{\emph{Proceedings of the Joint Workshop on Multimedia for Cooking and Eating Activities and Multimedia Assisted Dietary Management}}. \bibinfo{pages}{71--74}.
\newblock


\bibitem[Liu et~al\mbox{.}(2021)]%
        {9179998}
\bibfield{author}{\bibinfo{person}{Chengxu Liu}, \bibinfo{person}{Yuanzhi Liang}, \bibinfo{person}{Yao Xue}, \bibinfo{person}{Xueming Qian}, {and} \bibinfo{person}{Jianlong Fu}.} \bibinfo{year}{2021}\natexlab{}.
\newblock \showarticletitle{Food and Ingredient Joint Learning for Fine-Grained Recognition}.
\newblock \bibinfo{journal}{\emph{IEEE Transactions on Circuits and Systems for Video Technology}} \bibinfo{volume}{31}, \bibinfo{number}{6} (\bibinfo{year}{2021}), \bibinfo{pages}{2480--2493}.
\newblock
\urldef\tempurl%
\url{https://doi.org/10.1109/TCSVT.2020.3020079}
\showDOI{\tempurl}


\bibitem[Liu et~al\mbox{.}(2024)]%
        {liu2024canteen}
\bibfield{author}{\bibinfo{person}{Guoshan Liu}, \bibinfo{person}{Yang Jiao}, \bibinfo{person}{Jingjing Chen}, \bibinfo{person}{Bin Zhu}, {and} \bibinfo{person}{Yu-Gang Jiang}.} \bibinfo{year}{2024}\natexlab{}.
\newblock \showarticletitle{From Canteen Food to Daily Meals: Generalizing Food Recognition to More Practical Scenarios}.
\newblock \bibinfo{journal}{\emph{IEEE Transactions on Multimedia}} (\bibinfo{year}{2024}).
\newblock


\bibitem[Liu et~al\mbox{.}(2025)]%
        {liu2024retrieval}
\bibfield{author}{\bibinfo{person}{Guoshan Liu}, \bibinfo{person}{Hailong Yin}, \bibinfo{person}{Bin Zhu}, \bibinfo{person}{Jingjing Chen}, \bibinfo{person}{Chong-Wah Ngo}, {and} \bibinfo{person}{Yu-Gang Jiang}.} \bibinfo{year}{2025}\natexlab{}.
\newblock \showarticletitle{Retrieval Augmented Recipe Generation}. In \bibinfo{booktitle}{\emph{Proceedings of the IEEE/CVF Winter Conference on Applications of Computer Vision}}.
\newblock


\bibitem[Lu et~al\mbox{.}(2022)]%
        {lu2022dpm}
\bibfield{author}{\bibinfo{person}{Cheng Lu}, \bibinfo{person}{Yuhao Zhou}, \bibinfo{person}{Fan Bao}, \bibinfo{person}{Jianfei Chen}, \bibinfo{person}{Chongxuan Li}, {and} \bibinfo{person}{Jun Zhu}.} \bibinfo{year}{2022}\natexlab{}.
\newblock \showarticletitle{Dpm-solver: A fast ode solver for diffusion probabilistic model sampling in around 10 steps}.
\newblock \bibinfo{journal}{\emph{Advances in Neural Information Processing Systems}}  \bibinfo{volume}{35} (\bibinfo{year}{2022}), \bibinfo{pages}{5775--5787}.
\newblock


\bibitem[Lu et~al\mbox{.}(2023b)]%
        {lu2023dpmsolver}
\bibfield{author}{\bibinfo{person}{Cheng Lu}, \bibinfo{person}{Yuhao Zhou}, \bibinfo{person}{Fan Bao}, \bibinfo{person}{Jianfei Chen}, \bibinfo{person}{Chongxuan Li}, {and} \bibinfo{person}{Jun Zhu}.} \bibinfo{year}{2023}\natexlab{b}.
\newblock \bibinfo{title}{DPM-Solver++: Fast Solver for Guided Sampling of Diffusion Probabilistic Models}.
\newblock
\newblock
\showeprint[arxiv]{2211.01095}~[cs.LG]


\bibitem[Lu et~al\mbox{.}(2023a)]%
        {lu2023multimodal}
\bibfield{author}{\bibinfo{person}{Yujie Lu}, \bibinfo{person}{Pan Lu}, \bibinfo{person}{Zhiyu Chen}, \bibinfo{person}{Wanrong Zhu}, \bibinfo{person}{Xin~Eric Wang}, {and} \bibinfo{person}{William~Yang Wang}.} \bibinfo{year}{2023}\natexlab{a}.
\newblock \showarticletitle{Multimodal procedural planning via dual text-image prompting}.
\newblock \bibinfo{journal}{\emph{arXiv preprint arXiv:2305.01795}} (\bibinfo{year}{2023}).
\newblock


\bibitem[Mansimov et~al\mbox{.}(2015)]%
        {mansimov2015generating}
\bibfield{author}{\bibinfo{person}{Elman Mansimov}, \bibinfo{person}{Emilio Parisotto}, \bibinfo{person}{Jimmy~Lei Ba}, {and} \bibinfo{person}{Ruslan Salakhutdinov}.} \bibinfo{year}{2015}\natexlab{}.
\newblock \showarticletitle{Generating images from captions with attention}.
\newblock \bibinfo{journal}{\emph{arXiv preprint arXiv:1511.02793}} (\bibinfo{year}{2015}).
\newblock


\bibitem[Min et~al\mbox{.}(2019)]%
        {min2019survey}
\bibfield{author}{\bibinfo{person}{Weiqing Min}, \bibinfo{person}{Shuqiang Jiang}, \bibinfo{person}{Linhu Liu}, \bibinfo{person}{Yong Rui}, {and} \bibinfo{person}{Ramesh Jain}.} \bibinfo{year}{2019}\natexlab{}.
\newblock \showarticletitle{A survey on food computing}.
\newblock \bibinfo{journal}{\emph{ACM Computing Surveys (CSUR)}} \bibinfo{volume}{52}, \bibinfo{number}{5} (\bibinfo{year}{2019}), \bibinfo{pages}{1--36}.
\newblock


\bibitem[Min et~al\mbox{.}(2016)]%
        {min2016being}
\bibfield{author}{\bibinfo{person}{Weiqing Min}, \bibinfo{person}{Shuqiang Jiang}, \bibinfo{person}{Jitao Sang}, \bibinfo{person}{Huayang Wang}, \bibinfo{person}{Xinda Liu}, {and} \bibinfo{person}{Luis Herranz}.} \bibinfo{year}{2016}\natexlab{}.
\newblock \showarticletitle{Being a supercook: Joint food attributes and multimodal content modeling for recipe retrieval and exploration}.
\newblock \bibinfo{journal}{\emph{IEEE transactions on multimedia}} \bibinfo{volume}{19}, \bibinfo{number}{5} (\bibinfo{year}{2016}), \bibinfo{pages}{1100--1113}.
\newblock


\bibitem[Min et~al\mbox{.}(2023)]%
        {min2023large}
\bibfield{author}{\bibinfo{person}{Weiqing Min}, \bibinfo{person}{Zhiling Wang}, \bibinfo{person}{Yuxin Liu}, \bibinfo{person}{Mengjiang Luo}, \bibinfo{person}{Liping Kang}, \bibinfo{person}{Xiaoming Wei}, \bibinfo{person}{Xiaolin Wei}, {and} \bibinfo{person}{Shuqiang Jiang}.} \bibinfo{year}{2023}\natexlab{}.
\newblock \showarticletitle{Large scale visual food recognition}.
\newblock \bibinfo{journal}{\emph{IEEE Transactions on Pattern Analysis and Machine Intelligence}} (\bibinfo{year}{2023}).
\newblock


\bibitem[Ming et~al\mbox{.}(2018)]%
        {ming2018food}
\bibfield{author}{\bibinfo{person}{Zhao-Yan Ming}, \bibinfo{person}{Jingjing Chen}, \bibinfo{person}{Yu Cao}, \bibinfo{person}{Ciar{\'a}n Forde}, \bibinfo{person}{Chong-Wah Ngo}, {and} \bibinfo{person}{Tat~Seng Chua}.} \bibinfo{year}{2018}\natexlab{}.
\newblock \showarticletitle{Food photo recognition for dietary tracking: System and experiment}. In \bibinfo{booktitle}{\emph{MultiMedia Modeling: 24th International Conference, MMM 2018, Bangkok, Thailand, February 5-7, 2018, Proceedings, Part II 24}}. Springer, \bibinfo{pages}{129--141}.
\newblock


\bibitem[Mirza and Osindero(2014)]%
        {mirza2014conditional}
\bibfield{author}{\bibinfo{person}{Mehdi Mirza} {and} \bibinfo{person}{Simon Osindero}.} \bibinfo{year}{2014}\natexlab{}.
\newblock \showarticletitle{Conditional generative adversarial nets}.
\newblock \bibinfo{journal}{\emph{arXiv preprint arXiv:1411.1784}} (\bibinfo{year}{2014}).
\newblock


\bibitem[Qiu et~al\mbox{.}(2022)]%
        {qiu2022mining}
\bibfield{author}{\bibinfo{person}{Jianing Qiu}, \bibinfo{person}{Frank P-W Lo}, \bibinfo{person}{Yingnan Sun}, \bibinfo{person}{Siyao Wang}, {and} \bibinfo{person}{Benny Lo}.} \bibinfo{year}{2022}\natexlab{}.
\newblock \showarticletitle{Mining discriminative food regions for accurate food recognition}.
\newblock \bibinfo{journal}{\emph{arXiv preprint arXiv:2207.03692}} (\bibinfo{year}{2022}).
\newblock


\bibitem[Radford et~al\mbox{.}(2021)]%
        {radford2021learning}
\bibfield{author}{\bibinfo{person}{Alec Radford}, \bibinfo{person}{Jong~Wook Kim}, \bibinfo{person}{Chris Hallacy}, \bibinfo{person}{Aditya Ramesh}, \bibinfo{person}{Gabriel Goh}, \bibinfo{person}{Sandhini Agarwal}, \bibinfo{person}{Girish Sastry}, \bibinfo{person}{Amanda Askell}, \bibinfo{person}{Pamela Mishkin}, \bibinfo{person}{Jack Clark}, {et~al\mbox{.}}} \bibinfo{year}{2021}\natexlab{}.
\newblock \showarticletitle{Learning transferable visual models from natural language supervision}. In \bibinfo{booktitle}{\emph{International Conference on Machine Learning}}. PMLR, \bibinfo{pages}{8748--8763}.
\newblock


\bibitem[Reed et~al\mbox{.}(2016b)]%
        {reed2016generative}
\bibfield{author}{\bibinfo{person}{Scott Reed}, \bibinfo{person}{Zeynep Akata}, \bibinfo{person}{Xinchen Yan}, \bibinfo{person}{Lajanugen Logeswaran}, \bibinfo{person}{Bernt Schiele}, {and} \bibinfo{person}{Honglak Lee}.} \bibinfo{year}{2016}\natexlab{b}.
\newblock \showarticletitle{Generative adversarial text to image synthesis}. In \bibinfo{booktitle}{\emph{International Conference on Machine Learning}}. PMLR, \bibinfo{pages}{1060--1069}.
\newblock


\bibitem[Reed et~al\mbox{.}(2016c)]%
        {reed2016generating2}
\bibfield{author}{\bibinfo{person}{Scott Reed}, \bibinfo{person}{A{\"a}ron van~den Oord}, \bibinfo{person}{Nal Kalchbrenner}, \bibinfo{person}{Victor Bapst}, \bibinfo{person}{Matt Botvinick}, {and} \bibinfo{person}{Nando De~Freitas}.} \bibinfo{year}{2016}\natexlab{c}.
\newblock \showarticletitle{Generating interpretable images with controllable structure}.
\newblock  (\bibinfo{year}{2016}).
\newblock


\bibitem[Reed et~al\mbox{.}(2016a)]%
        {reed2016learning}
\bibfield{author}{\bibinfo{person}{Scott~E Reed}, \bibinfo{person}{Zeynep Akata}, \bibinfo{person}{Santosh Mohan}, \bibinfo{person}{Samuel Tenka}, \bibinfo{person}{Bernt Schiele}, {and} \bibinfo{person}{Honglak Lee}.} \bibinfo{year}{2016}\natexlab{a}.
\newblock \showarticletitle{Learning what and where to draw}.
\newblock \bibinfo{journal}{\emph{Advances in Neural Information Processing Systems}}  \bibinfo{volume}{29} (\bibinfo{year}{2016}).
\newblock


\bibitem[Rombach et~al\mbox{.}(2022)]%
        {rombach2022high}
\bibfield{author}{\bibinfo{person}{Robin Rombach}, \bibinfo{person}{Andreas Blattmann}, \bibinfo{person}{Dominik Lorenz}, \bibinfo{person}{Patrick Esser}, {and} \bibinfo{person}{Bj{\"o}rn Ommer}.} \bibinfo{year}{2022}\natexlab{}.
\newblock \showarticletitle{High-resolution image synthesis with latent diffusion models}. In \bibinfo{booktitle}{\emph{Proceedings of the IEEE/CVF Conference on Computer Vision and Pattern Recognition}}. \bibinfo{pages}{10684--10695}.
\newblock


\bibitem[Sahoo et~al\mbox{.}(2019)]%
        {sahoo2019foodai}
\bibfield{author}{\bibinfo{person}{Doyen Sahoo}, \bibinfo{person}{Wang Hao}, \bibinfo{person}{Shu Ke}, \bibinfo{person}{Wu Xiongwei}, \bibinfo{person}{Hung Le}, \bibinfo{person}{Palakorn Achananuparp}, \bibinfo{person}{Ee-Peng Lim}, {and} \bibinfo{person}{Steven~CH Hoi}.} \bibinfo{year}{2019}\natexlab{}.
\newblock \showarticletitle{FoodAI: Food image recognition via deep learning for smart food logging}. In \bibinfo{booktitle}{\emph{Proceedings of the 25th ACM SIGKDD International Conference on Knowledge Discovery \& Data Mining}}. \bibinfo{pages}{2260--2268}.
\newblock


\bibitem[Salvador et~al\mbox{.}(2019)]%
        {salvador2019inverse}
\bibfield{author}{\bibinfo{person}{Amaia Salvador}, \bibinfo{person}{Michal Drozdzal}, \bibinfo{person}{Xavier Gir{\'o}-i Nieto}, {and} \bibinfo{person}{Adriana Romero}.} \bibinfo{year}{2019}\natexlab{}.
\newblock \showarticletitle{Inverse cooking: Recipe generation from food images}. In \bibinfo{booktitle}{\emph{Proceedings of the IEEE/CVF Conference on Computer Vision and Pattern Recognition}}. \bibinfo{pages}{10453--10462}.
\newblock


\bibitem[Salvador et~al\mbox{.}(2021)]%
        {salvador2021revamping}
\bibfield{author}{\bibinfo{person}{Amaia Salvador}, \bibinfo{person}{Erhan Gundogdu}, \bibinfo{person}{Loris Bazzani}, {and} \bibinfo{person}{Michael Donoser}.} \bibinfo{year}{2021}\natexlab{}.
\newblock \showarticletitle{Revamping cross-modal recipe retrieval with hierarchical transformers and self-supervised learning}. In \bibinfo{booktitle}{\emph{Proceedings of the IEEE/CVF Conference on Computer Vision and Pattern Recognition}}. \bibinfo{pages}{15475--15484}.
\newblock


\bibitem[Salvador et~al\mbox{.}(2017)]%
        {salvador2017learning}
\bibfield{author}{\bibinfo{person}{Amaia Salvador}, \bibinfo{person}{Nicholas Hynes}, \bibinfo{person}{Yusuf Aytar}, \bibinfo{person}{Javier Marin}, \bibinfo{person}{Ferda Ofli}, \bibinfo{person}{Ingmar Weber}, {and} \bibinfo{person}{Antonio Torralba}.} \bibinfo{year}{2017}\natexlab{}.
\newblock \showarticletitle{Learning cross-modal embeddings for cooking recipes and food images}. In \bibinfo{booktitle}{\emph{Proceedings of the IEEE Conference on Computer Vision and Pattern Recognition}}. \bibinfo{pages}{3020--3028}.
\newblock


\bibitem[Shen et~al\mbox{.}(2022)]%
        {shen2021much}
\bibfield{author}{\bibinfo{person}{Sheng Shen}, \bibinfo{person}{Liunian~Harold Li}, \bibinfo{person}{Hao Tan}, \bibinfo{person}{Mohit Bansal}, \bibinfo{person}{Anna Rohrbach}, \bibinfo{person}{Kai-Wei Chang}, \bibinfo{person}{Zhewei Yao}, {and} \bibinfo{person}{Kurt Keutzer}.} \bibinfo{year}{2022}\natexlab{}.
\newblock \showarticletitle{How much can clip benefit vision-and-language tasks?}. In \bibinfo{booktitle}{\emph{The Tenth International Conference on Learning Representations}}.
\newblock


\bibitem[Skorokhodov et~al\mbox{.}(2022)]%
        {skorokhodov2022stylegan}
\bibfield{author}{\bibinfo{person}{Ivan Skorokhodov}, \bibinfo{person}{Sergey Tulyakov}, {and} \bibinfo{person}{Mohamed Elhoseiny}.} \bibinfo{year}{2022}\natexlab{}.
\newblock \showarticletitle{Stylegan-v: A continuous video generator with the price, image quality and perks of stylegan2}. In \bibinfo{booktitle}{\emph{Proceedings of the IEEE/CVF Conference on Computer Vision and Pattern Recognition}}. \bibinfo{pages}{3626--3636}.
\newblock


\bibitem[Song et~al\mbox{.}(2025)]%
        {song2025enhancing}
\bibfield{author}{\bibinfo{person}{Fangzhou Song}, \bibinfo{person}{Bin Zhu}, \bibinfo{person}{Yanbin Hao}, {and} \bibinfo{person}{Shuo Wang}.} \bibinfo{year}{2025}\natexlab{}.
\newblock \showarticletitle{Enhancing recipe retrieval with foundation models: A data augmentation perspective}. In \bibinfo{booktitle}{\emph{European Conference on Computer Vision}}. Springer, \bibinfo{pages}{111--127}.
\newblock


\bibitem[Song et~al\mbox{.}(2021)]%
        {song2020denoising}
\bibfield{author}{\bibinfo{person}{Jiaming Song}, \bibinfo{person}{Chenlin Meng}, {and} \bibinfo{person}{Stefano Ermon}.} \bibinfo{year}{2021}\natexlab{}.
\newblock \showarticletitle{Denoising diffusion implicit models}. In \bibinfo{booktitle}{\emph{9th International Conference on Learning Representations}}. \bibinfo{publisher}{OpenReview.net}.
\newblock


\bibitem[Szegedy et~al\mbox{.}(2016)]%
        {szegedy2016rethinking}
\bibfield{author}{\bibinfo{person}{Christian Szegedy}, \bibinfo{person}{Vincent Vanhoucke}, \bibinfo{person}{Sergey Ioffe}, \bibinfo{person}{Jon Shlens}, {and} \bibinfo{person}{Zbigniew Wojna}.} \bibinfo{year}{2016}\natexlab{}.
\newblock \showarticletitle{Rethinking the inception architecture for computer vision}. In \bibinfo{booktitle}{\emph{Proceedings of the IEEE Conference on Computer Vision and Pattern Recognition}}. \bibinfo{pages}{2818--2826}.
\newblock


\bibitem[Tao et~al\mbox{.}(2022)]%
        {tao2022df}
\bibfield{author}{\bibinfo{person}{Ming Tao}, \bibinfo{person}{Hao Tang}, \bibinfo{person}{Fei Wu}, \bibinfo{person}{Xiao-Yuan Jing}, \bibinfo{person}{Bing-Kun Bao}, {and} \bibinfo{person}{Changsheng Xu}.} \bibinfo{year}{2022}\natexlab{}.
\newblock \showarticletitle{Df-gan: A simple and effective baseline for text-to-image synthesis}. In \bibinfo{booktitle}{\emph{Proceedings of the IEEE/CVF Conference on Computer Vision and Pattern Recognition}}. \bibinfo{pages}{16515--16525}.
\newblock


\bibitem[Van Den~Oord et~al\mbox{.}(2017)]%
        {van2017neural}
\bibfield{author}{\bibinfo{person}{Aaron Van Den~Oord}, \bibinfo{person}{Oriol Vinyals}, {et~al\mbox{.}}} \bibinfo{year}{2017}\natexlab{}.
\newblock \showarticletitle{Neural discrete representation learning}.
\newblock \bibinfo{journal}{\emph{Advances in Neural Information Processing Systems}}  \bibinfo{volume}{30} (\bibinfo{year}{2017}).
\newblock


\bibitem[Villegas et~al\mbox{.}(2022)]%
        {villegas2022phenaki}
\bibfield{author}{\bibinfo{person}{Ruben Villegas}, \bibinfo{person}{Mohammad Babaeizadeh}, \bibinfo{person}{Pieter-Jan Kindermans}, \bibinfo{person}{Hernan Moraldo}, \bibinfo{person}{Han Zhang}, \bibinfo{person}{Mohammad~Taghi Saffar}, \bibinfo{person}{Santiago Castro}, \bibinfo{person}{Julius Kunze}, {and} \bibinfo{person}{Dumitru Erhan}.} \bibinfo{year}{2022}\natexlab{}.
\newblock \showarticletitle{Phenaki: Variable length video generation from open domain textual description}.
\newblock \bibinfo{journal}{\emph{arXiv preprint arXiv:2210.02399}} (\bibinfo{year}{2022}).
\newblock


\bibitem[Wang et~al\mbox{.}(2021)]%
        {wang2021cross}
\bibfield{author}{\bibinfo{person}{Hao Wang}, \bibinfo{person}{Doyen Sahoo}, \bibinfo{person}{Chenghao Liu}, \bibinfo{person}{Ke Shu}, \bibinfo{person}{Palakorn Achananuparp}, \bibinfo{person}{Ee-peng Lim}, {and} \bibinfo{person}{Steven~CH Hoi}.} \bibinfo{year}{2021}\natexlab{}.
\newblock \showarticletitle{Cross-modal food retrieval: learning a joint embedding of food images and recipes with semantic consistency and attention mechanism}.
\newblock \bibinfo{journal}{\emph{IEEE Transactions on Multimedia}}  \bibinfo{volume}{24} (\bibinfo{year}{2021}), \bibinfo{pages}{2515--2525}.
\newblock


\bibitem[Xu et~al\mbox{.}(2018)]%
        {xu2018attngan}
\bibfield{author}{\bibinfo{person}{Tao Xu}, \bibinfo{person}{Pengchuan Zhang}, \bibinfo{person}{Qiuyuan Huang}, \bibinfo{person}{Han Zhang}, \bibinfo{person}{Zhe Gan}, \bibinfo{person}{Xiaolei Huang}, {and} \bibinfo{person}{Xiaodong He}.} \bibinfo{year}{2018}\natexlab{}.
\newblock \showarticletitle{Attngan: Fine-grained text to image generation with attentional generative adversarial networks}. In \bibinfo{booktitle}{\emph{Proceedings of the IEEE Conference on Computer Vision and Pattern Recognition}}. \bibinfo{pages}{1316--1324}.
\newblock


\bibitem[Yin et~al\mbox{.}(2023b)]%
        {yin2023nuwa}
\bibfield{author}{\bibinfo{person}{Shengming Yin}, \bibinfo{person}{Chenfei Wu}, \bibinfo{person}{Huan Yang}, \bibinfo{person}{Jianfeng Wang}, \bibinfo{person}{Xiaodong Wang}, \bibinfo{person}{Minheng Ni}, \bibinfo{person}{Zhengyuan Yang}, \bibinfo{person}{Linjie Li}, \bibinfo{person}{Shuguang Liu}, \bibinfo{person}{Fan Yang}, {et~al\mbox{.}}} \bibinfo{year}{2023}\natexlab{b}.
\newblock \showarticletitle{NUWA-XL: Diffusion over Diffusion for eXtremely Long Video Generation}.
\newblock \bibinfo{journal}{\emph{arXiv preprint arXiv:2303.12346}} (\bibinfo{year}{2023}).
\newblock


\bibitem[Yin et~al\mbox{.}(2023a)]%
        {yin2023foodlmm}
\bibfield{author}{\bibinfo{person}{Yuehao Yin}, \bibinfo{person}{Huiyan Qi}, \bibinfo{person}{Bin Zhu}, \bibinfo{person}{Jingjing Chen}, \bibinfo{person}{Yu-Gang Jiang}, {and} \bibinfo{person}{Chong-Wah Ngo}.} \bibinfo{year}{2023}\natexlab{a}.
\newblock \showarticletitle{Foodlmm: A versatile food assistant using large multi-modal model}.
\newblock \bibinfo{journal}{\emph{arXiv preprint arXiv:2312.14991}} (\bibinfo{year}{2023}).
\newblock


\bibitem[Yu et~al\mbox{.}(2023)]%
        {yu2023video}
\bibfield{author}{\bibinfo{person}{Sihyun Yu}, \bibinfo{person}{Kihyuk Sohn}, \bibinfo{person}{Subin Kim}, {and} \bibinfo{person}{Jinwoo Shin}.} \bibinfo{year}{2023}\natexlab{}.
\newblock \showarticletitle{Video probabilistic diffusion models in projected latent space}. In \bibinfo{booktitle}{\emph{Proceedings of the IEEE/CVF Conference on Computer Vision and Pattern Recognition}}. \bibinfo{pages}{18456--18466}.
\newblock


\bibitem[Zhang et~al\mbox{.}(2021)]%
        {zhang2021cross}
\bibfield{author}{\bibinfo{person}{Han Zhang}, \bibinfo{person}{Jing~Yu Koh}, \bibinfo{person}{Jason Baldridge}, \bibinfo{person}{Honglak Lee}, {and} \bibinfo{person}{Yinfei Yang}.} \bibinfo{year}{2021}\natexlab{}.
\newblock \showarticletitle{Cross-modal contrastive learning for text-to-image generation}. In \bibinfo{booktitle}{\emph{Proceedings of the IEEE/CVF Conference on Computer Vision and Pattern Recognition}}. \bibinfo{pages}{833--842}.
\newblock


\bibitem[Zhang et~al\mbox{.}(2017)]%
        {zhang2017stackgan}
\bibfield{author}{\bibinfo{person}{Han Zhang}, \bibinfo{person}{Tao Xu}, \bibinfo{person}{Hongsheng Li}, \bibinfo{person}{Shaoting Zhang}, \bibinfo{person}{Xiaogang Wang}, \bibinfo{person}{Xiaolei Huang}, {and} \bibinfo{person}{Dimitris~N Metaxas}.} \bibinfo{year}{2017}\natexlab{}.
\newblock \showarticletitle{Stackgan: Text to photo-realistic image synthesis with stacked generative adversarial networks}. In \bibinfo{booktitle}{\emph{Proceedings of the IEEE International Conference on Computer Vision}}. \bibinfo{pages}{5907--5915}.
\newblock


\bibitem[Zhang et~al\mbox{.}(2018)]%
        {zhang2018stackgan++}
\bibfield{author}{\bibinfo{person}{Han Zhang}, \bibinfo{person}{Tao Xu}, \bibinfo{person}{Hongsheng Li}, \bibinfo{person}{Shaoting Zhang}, \bibinfo{person}{Xiaogang Wang}, \bibinfo{person}{Xiaolei Huang}, {and} \bibinfo{person}{Dimitris~N Metaxas}.} \bibinfo{year}{2018}\natexlab{}.
\newblock \showarticletitle{Stackgan++: Realistic image synthesis with stacked generative adversarial networks}.
\newblock \bibinfo{journal}{\emph{IEEE Transactions on Pattern Analysis and Machine Intelligence}} \bibinfo{volume}{41}, \bibinfo{number}{8} (\bibinfo{year}{2018}), \bibinfo{pages}{1947--1962}.
\newblock


\bibitem[Zhang et~al\mbox{.}(2023)]%
        {zhang2023adding}
\bibfield{author}{\bibinfo{person}{Lvmin Zhang}, \bibinfo{person}{Anyi Rao}, {and} \bibinfo{person}{Maneesh Agrawala}.} \bibinfo{year}{2023}\natexlab{}.
\newblock \showarticletitle{Adding conditional control to text-to-image diffusion models}. In \bibinfo{booktitle}{\emph{Proceedings of the IEEE/CVF International Conference on Computer Vision}}. \bibinfo{pages}{3836--3847}.
\newblock


\bibitem[Zhao et~al\mbox{.}(2021)]%
        {zhao2021fusion}
\bibfield{author}{\bibinfo{person}{Heng Zhao}, \bibinfo{person}{Kim-Hui Yap}, {and} \bibinfo{person}{Alex~Chichung Kot}.} \bibinfo{year}{2021}\natexlab{}.
\newblock \showarticletitle{Fusion learning using semantics and graph convolutional network for visual food recognition}. In \bibinfo{booktitle}{\emph{Proceedings of the IEEE/CVF Winter Conference on Applications of Computer Vision}}. \bibinfo{pages}{1711--1720}.
\newblock


\bibitem[Zhao et~al\mbox{.}(2023)]%
        {zhao2023multi}
\bibfield{author}{\bibinfo{person}{Liang Zhao}, \bibinfo{person}{Pingda Huang}, \bibinfo{person}{Tengtuo Chen}, \bibinfo{person}{Chunjiang Fu}, \bibinfo{person}{Qinghao Hu}, {and} \bibinfo{person}{Yangqianhui Zhang}.} \bibinfo{year}{2023}\natexlab{}.
\newblock \showarticletitle{Multi-Sentence Complementarily Generation for Text-to-Image Synthesis}.
\newblock \bibinfo{journal}{\emph{IEEE Transactions on Multimedia}} (\bibinfo{year}{2023}).
\newblock


\bibitem[Zhou et~al\mbox{.}(2018)]%
        {zhou2018towards}
\bibfield{author}{\bibinfo{person}{Luowei Zhou}, \bibinfo{person}{Chenliang Xu}, {and} \bibinfo{person}{Jason Corso}.} \bibinfo{year}{2018}\natexlab{}.
\newblock \showarticletitle{Towards automatic learning of procedures from web instructional videos}. In \bibinfo{booktitle}{\emph{Proceedings of the AAAI Conference on Artificial Intelligence}}, Vol.~\bibinfo{volume}{32}.
\newblock


\bibitem[Zhu and Ngo(2020)]%
        {zhu2020cookgan}
\bibfield{author}{\bibinfo{person}{Bin Zhu} {and} \bibinfo{person}{Chong-Wah Ngo}.} \bibinfo{year}{2020}\natexlab{}.
\newblock \showarticletitle{CookGAN: Causality based text-to-image synthesis}. In \bibinfo{booktitle}{\emph{Proceedings of the IEEE/CVF Conference on Computer Vision and Pattern Recognition}}. \bibinfo{pages}{5519--5527}.
\newblock


\bibitem[Zhu et~al\mbox{.}(2021)]%
        {zhu2021learning}
\bibfield{author}{\bibinfo{person}{Bin Zhu}, \bibinfo{person}{Chong-Wah Ngo}, {and} \bibinfo{person}{Wing-Kwong Chan}.} \bibinfo{year}{2021}\natexlab{}.
\newblock \showarticletitle{Learning from web recipe-image pairs for food recognition: Problem, baselines and performance}.
\newblock \bibinfo{journal}{\emph{IEEE Transactions on Multimedia}}  \bibinfo{volume}{24} (\bibinfo{year}{2021}), \bibinfo{pages}{1175--1185}.
\newblock


\bibitem[Zhu et~al\mbox{.}(2019)]%
        {zhu2019r2gan}
\bibfield{author}{\bibinfo{person}{Bin Zhu}, \bibinfo{person}{Chong-Wah Ngo}, \bibinfo{person}{Jingjing Chen}, {and} \bibinfo{person}{Yanbin Hao}.} \bibinfo{year}{2019}\natexlab{}.
\newblock \showarticletitle{R2gan: Cross-modal recipe retrieval with generative adversarial network}. In \bibinfo{booktitle}{\emph{Proceedings of the IEEE/CVF Conference on Computer Vision and Pattern Recognition}}. \bibinfo{pages}{11477--11486}.
\newblock


\end{thebibliography}
\end{document}